\documentclass[lettersize,journal]{IEEEtran}
\usepackage{amsmath,amsfonts}
\usepackage{algorithmic}
\usepackage{array}
\usepackage[caption=false,font=normalsize,labelfont=sf,textfont=sf]{subfig}
\usepackage{textcomp}
\usepackage{stfloats}
\usepackage{url}
\usepackage{verbatim}
\usepackage{graphicx}
\usepackage{hyperref}
\usepackage{url}  
\usepackage{hyperref}
\usepackage{xcolor}

\hypersetup{
    colorlinks=true,
    citecolor=cyan,    
    linkcolor=cyan,     
    urlcolor=cyan
}

\usepackage{subcaption}
\usepackage{subfig}
\usepackage{algorithm}
\usepackage{algorithmic}
\usepackage{epsfig}
\usepackage{graphicx}
\usepackage{amsmath}
\usepackage{amssymb}
\usepackage[table]{xcolor} 
\usepackage{array}         
\usepackage{booktabs}
\usepackage{tikz}
\usepackage{tikz}
\usepackage{xcolor} 
\usepackage{multirow}
\usepackage{comment}
\usepackage{lineno}
\usepackage[font=small,labelfont=normalfont,textfont=normalfont]
{caption}
\definecolor{c1}{rgb}{0.435, 0.584, 0.933}
\definecolor{c2}{rgb}{0.533, 0.890, 0.949}
\definecolor{c3}{rgb}{0.145, 0.231, 0.573}
\definecolor{c4}{rgb}{0.290, 0.125, 0.675}
\definecolor{c5}{rgb}{0.376, 0.318, 0.949}
\definecolor{c6}{rgb}{0.917, 0.235, 0.192}
\definecolor{c7}{rgb}{0.917, 0.259, 0.765}
\definecolor{c8}{rgb}{0.541, 0.165, 0.353}
\definecolor{c9}{rgb}{0.917, 0.200, 0.964}
\definecolor{c10}{rgb}{0.945, 0.612, 0.976}
\definecolor{c11}{rgb}{0.267, 0.031, 0.282}
\definecolor{c12}{rgb}{0.627, 0.125, 0.298}
\definecolor{c13}{rgb}{0.961, 0.792, 0.271}
\definecolor{c14}{rgb}{0.933, 0.502, 0.271}
\definecolor{c15}{rgb}{0.310, 0.678, 0.196}
\definecolor{c16}{rgb}{0.494, 0.255, 0.086}
\definecolor{c17}{rgb}{0.671, 0.933, 0.412}
\definecolor{c18}{rgb}{0.988, 0.945, 0.631}
\definecolor{c19}{rgb}{0.914, 0.200, 0.137}
\definecolor{c20}{rgb}{0.980, 0.588, 0.000}
\definecolor{c21}{rgb}{0.196, 1.000, 1.000}

\usepackage{tikz,xcolor,hyperref}
\definecolor{lime}{HTML}{A6CE39}
\DeclareRobustCommand{\orcidicon}{%
    \begin{tikzpicture}
    \draw[lime, fill=lime] (0,0)
    circle [radius=0.16]
    node[white] {{\fontfamily{qag}\selectfont \tiny ID}};    \draw[white, fill=white] (-0.0625,0.095)
    circle [radius=0.007];    \end{tikzpicture}
    \hspace{-2mm}}
\foreach \x in {A, ..., Z}{%
    \expandafter\xdef\csname orcid\x\endcsname{\noexpand\href{https://orcid.org/\csname orcidauthor\x\endcsname}{\noexpand\orcidicon}}
    }

\def\BibTeX{{\rm B\kern-.05em{\sc i\kern-.025em b}\kern-.08em
    T\kern-.1667em\lower.7ex\hbox{E}\kern-.125emX}}
\usepackage{balance}
\begin{document}

\title{DSOcc: Leveraging Depth Awareness and Semantic Aid to Boost Camera-Based 3D Semantic Occupancy Prediction}

\newcommand{\orcidauthorA}{0000-0003-0145-1690}
\newcommand{\orcidauthorB}{0000-0002-0365-6166}
\newcommand{\orcidauthorC}{0000-0001-5745-5530}
\newcommand{\orcidauthorD}{0000-0002-5690-8275}
\newcommand{\orcidauthorG}{0000-0002-0597-4475}
\newcommand{\orcidauthorH}{0000-0002-0329-7458}
\newcommand{\orcidauthorF}{0000-0001-9023-5361}
\newcommand{\orcidauthorE}{0000-0001-9358-0099}

\author{Naiyu Fang\orcidA{}, Zheyuan Zhou\orcidB{}, Kang Wang\orcidC{}, Ruibo Li\orcidD{}, Lemiao Qiu\orcidE{}, Shuyou Zhang\orcidF{}, Zhe Wang\orcidG{}, Guosheng Lin\orcidH{}
\thanks{\it Corresponding authors: Guosheng Lin}
\thanks{Naiyu Fang is with S-Lab \& College of Computing and Data Science, Nanyang Technological University, Singapore, 
and also with MMLab, The Chinese University of Hong Kong, Hong Kong 
(e-mail: FangNaiyu@zju.edu.cn; naiyufang@cuhk.edu.hk).}%
\thanks{Ruibo Li and Guosheng Lin are with S-Lab \& College of Computing and Data Science, Nanyang Technological University, Singapore (e-mail: ruibo.li@ntu.edu.sg; gslin@ntu.edu.sg).}%
\thanks{Zheyuan Zhou, Kang Wang, Lemiao Qiu, and Shuyou Zhang are with the State Key Laboratory of Fluid Power \& Mechatronic Systems, Zhejiang University, Hangzhou, China (e-mail: zheyuanzhou@zju.edu.cn; kang.wang@zju.edu.cn; qiulm@zju.edu.cn; zsy@zju.edu.cn).}%
\thanks{Zhe Wang is with SenseTime Research, Hong Kong (e-mail: wangzhe@sensetime.com).}%
}

\maketitle

\begin{abstract}
Camera-based 3D semantic occupancy prediction offers an efficient and cost-effective solution for perceiving surrounding scenes in autonomous driving. However, existing works rely on explicit occupancy state inference, leading to numerous incorrect feature assignments, and insufficient samples restrict the learning of occupancy class inference. To address these challenges, we propose leveraging \textbf{D}epth awareness and \textbf{S}emantic aid to boost camera-based 3D semantic \textbf{Occ}upancy prediction (\textbf{DSOcc}). We jointly perform occupancy state and occupancy class inference, where soft occupancy confidence is calculated by non-learning method and multiplied with image features to make voxels aware of depth, enabling adaptive implicit occupancy state inference. Instead of enhancing feature learning, we directly utilize well-trained image semantic segmentation and fuse multiple frames with their occupancy probabilities to aid occupancy class inference, thereby enhancing robustness. Experimental results demonstrate that DSOcc achieves state-of-the-art performance on the SemanticKITTI dataset among camera-based methods and achieves competitive performance on the SSCBench-KITTI-360 and Occ3D-nuScenes datasets. Code will be released on github.
\end{abstract}

\section{Introduction}
The perception of surrounding scenes serves as the foundational part of autonomous driving. Compared to the LiDAR-based scheme \cite{yang2024mcsgcalib,luo2019probability,mei2019semantic}, the camera-based scheme \cite{zhou2022cross,chen2016monocular} has attracted significant attention due to the lower cost of devices. To achieve 3D perception, camera-based 3D semantic occupancy prediction \cite{li2023voxformer,zhang2023occformer,xiao2024instance,ren2025rm,shi2025offboard} represents the surrounding environment using a predefined voxel grid, where each voxel is annotated with a semantic label. This task marks a significant advancement towards cheap, clear, efficient scene understanding for autonomous driving.

Camera-based 3D semantic occupancy prediction extracts features from camera images and projects them onto a voxel grid for prediction. This task requires consideration of two aspects: occupancy state and occupancy class. The occupancy state determines whether a voxel is occupied, while the occupancy class specifies which class a voxel belongs to. Typically, the model uses the projection matrix and depth information to predict the occupancy state, and then utilizes image features to predict the occupancy class.

\begin{figure}[t] 
\begin{center}
\includegraphics[width=\linewidth]{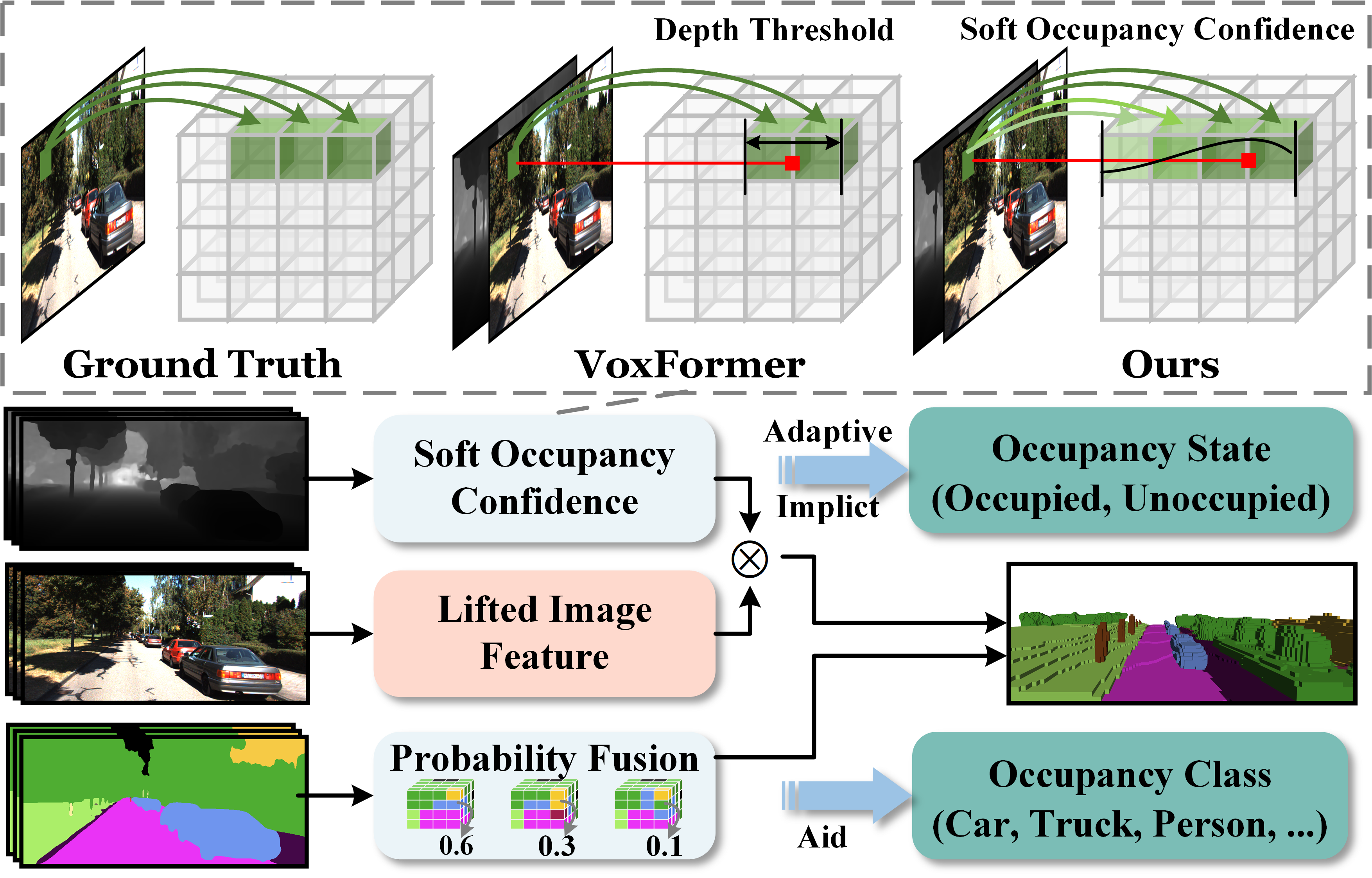}  
\end{center}
\caption{Camera-based 3D semantic occupancy prediction methods typically rely on depth to determine occupancy states. Depth-threshold-based approaches, however, are prone to misclassifying occupied voxels as empty, which leads to accumulated errors in subsequent occupancy class inference. In contrast, we employ a soft occupancy confidence to implicitly represent the occupancy state, effectively addressing this issue. Furthermore, we introduce probability-fused image semantic segmentation to assist occupancy class inference, enabling robust and accurate reasoning without increasing the scale of occupancy annotations. }
\label{fig1}
\end{figure}

For the occupancy state, two-stage methods \cite{li2023voxformer,mei2024camera} and depth-query methods \cite{jiang2024symphonize,wang2024h2gformer} estimate a depth threshold to classify voxels as either occupied or unoccupied based on depth and voxel position. The issue arises when depth prediction or threshold estimation is inaccurate: since only occupied voxels are assigned features for subsequent classification, a truly occupied voxel may be misclassified as unoccupied, while its neighboring unoccupied voxel may incorrectly receive features. \textbf{This phenomenon is common. Nearly 40\% of occupied voxels in \cite{li2023voxformer} are misclassified as unoccupied}. Specifically, we evaluate the Stage-1 module in Voxformer \cite{li2023voxformer} on the SemanticKITTI validation dataset \cite{behley2019semantickitti}. The occupancy inference results are depicted in Table \ref{tableA1}. While the result of ${{{FN} \mathord{\left/
 {\vphantom {{FN} {{N_a}}}} \right.
 \kern-\nulldelimiterspace} {{N_a}}}}$ appear promising, it is important to note that occupied voxels represent a small proportion of all voxels (${{{{N_v}} \mathord{\left/
 {\vphantom {{{N_v}} {{N_a}}}} \right.
 \kern-\nulldelimiterspace} {{N_a}}}}$ = 14.60\%). As a result, 38.14\% of occupied voxels are incorrectly inferred as unoccupied. These false negatives do not collect any image features in the subsequent stage, leading to numerous inference errors. To mitigate accumulated errors, it is crucial to reduce the heavy reliance on depth prediction or threshold estimation.

\begin{table}[!htb]
\centering
\caption{The occupancy status inference results for VoxFormer. ${FN}$, ${{N_a}}$, ${{N_v}}$ denote the number of false negative in prediction, the total number of voxels, and the number of occupied voxels in the ground truth, respectively.}
\begin{tabular}{llll}                                  
\toprule
Metric & ${{{FN} \mathord{\left/
 {\vphantom {{FN} {{N_a}}}} \right.
 \kern-\nulldelimiterspace} {{N_a}}}}$
& ${{{{N_v}} \mathord{\left/
 {\vphantom {{{N_v}} {{N_a}}}} \right.
 \kern-\nulldelimiterspace} {{N_a}}}}$
 & ${{{FN} \mathord{\left/
 {\vphantom {{FN} {{N_v}}}} \right.
 \kern-\nulldelimiterspace} {{N_v}}}}$
\\
\midrule
Value (\%) & 5.65 & 14.60& 38.14\\
\bottomrule
\end{tabular}
\label{tableA1}
\end{table}

For the occupancy class, the prediction accuracy relies on the mapping effect from image features to occupancy classes. To this end, previous methods have incorporated various modules to learn geometric features \cite{xue2024bi}, instance cues \cite{jiang2024symphonize}, and semantic guidance \cite{mei2024camera}. However, a critical issue overlooked is that they are trained on limited datasets. For instance, SemanticKITTI \cite{behley2019semantickitti} and SSCBenchKITTI-360 \cite{li2024sscbench} contain only 3,834 and 8,487 training samples, respectively. Within this fact, effectively utilizing image features is akin to 'dancing in shackles', and the occupancy class prediction are constrained in this shackle.

This naturally leads to the questions:  \textcolor{blue}{\textit{how can depth information be effectively exploited, and how can class reasoning limitations be broken, in camera-based semantic occupancy prediction?}}

To address these challenges, we propose depth awareness and semantic aid to boost camera-based 3D semantic occupancy prediction in the aspect of occupancy state and occupancy class. As illustrated in Fig. \ref{fig1}, We embed soft occupancy confidence into image feature to enable adaptive implicit occupancy state inference and integrate image semantic segmentation to enhance the robustness of occupancy class inference. Specifically, we perform occupancy state and occupancy class inference jointly, rather than treating them as independent subtasks. We directly calculate soft occupancy confidence by measuring the distance between voxel position and depth. The soft occupancy confidence is multiplied with the image features and assigned to the corresponding voxel. This prevents the loss of image features in false unoccupied voxel while enabling depth-aware image feature collection. We observe that 3D semantic occupancy prediction and image semantic segmentation share the same objective, while image semantic segmentation demonstrates better robustness when trained on large-scale datasets. Thus, we weight multi-frame semantic segmentation maps with occupancy probabilities to aid the occupancy class inference. Finally, we design a voxel fusion module to integrate the two voxel types for the final prediction. In summary, the main contributions of this work are summarized as follows:

\textbullet\ We propose calculating soft occupancy confidence in a non-learning method to embed depth into image features, enabling adaptive implicit occupancy state inference.

\textbullet\ We propose fusing multi-frame image semantic segmentation with occupancy probabilities to aid in occupancy class inference, thereby enhancing its robustness.

\textbullet\ Experiments show that our DSOcc achieves state-of-the-art performance, obtaining 18.02 and 16.90 mIoU on the SemanticKITTI validation and hidden test sets, respectively; 17.88 mIoU on the SSCBench-KITTI-360 validation set; and 28.40 mIoU on the Occ3D-nuScenes validation set.

\section{Related Works}

\textbf{3D Semantic Occupancy Prediction.} 3D semantic occupancy prediction originates from the semantic scene completion (SSC) task, first introduced in SSCNet \cite{zhudeformable}. SSC extracts images features, lifts them to a 3D representation, and then predicts voxels, each assigned a semantic label \cite{liu2021swin,song2017semantic,li2019rgbd,zhang2019cascaded, liu2018see}. As autonomous driving technology advances \cite{li2022modality,fang2024cross,li2022panoptic}, 3D semantic occupancy prediction has gained attention for providing clear and efficient representations. Among these, camera-based methods \cite{cao2022monoscene, huang2024selfocc,zhang2023occformer} are popular due to the cost-effectiveness of camera devices.

For camera-based 3D semantic occupancy prediction, dimensional transformation is a crucial component. MonoScene \cite{cao2022monoscene} firstly introduced a 2D-to-3D feature projection. OccFormer \cite{zhang2023occformer} leverages local and global transformers to perform projection along the horizontal plane. SurroundOcc \cite{wei2023surroundocc} employed spatial projection to lift multi-scale features. COTR \cite{ma2024cotr} introduced an explicit-implicit transformation for compact 3D representation. ViewFormer \cite{li2025viewformer} proposed a learning-first view attention mechanism to aggregate multi-view features.

Following dimensional transformation, previous methods utilize depth information to resolve projection ambiguity during feature collection. Two-stage method \cite{li2023voxformer,mei2024camera} learns an explicit occupancy status conditioned on depth and then collects features only for occupied voxels. The depth-query method \cite{jiang2024symphonize,wang2024h2gformer} uses depth as a query and interacts with features to embed depth information. Similar to LSS \cite{philion2020lift}, methods in \cite{zhao2024lowrankocc, zhang2023occformer} predict context and depth features from images and use the outer product to incorporate depth awareness. ViDAR \cite{yang2024visual} employs ray-casting renders to predict depth conditional probability and learns to multiply it with BEV features to embed depth information.

Inspired by \cite{philion2020lift,yang2024visual}, DSOcc aims to calculate soft occupancy confidence based on depth and embed it into image features for occupancy status inference. This is achieved through direct calculation without requiring any learning module.

\textbf{2D Image Representation.} Besides image features, camera-based 3D semantic occupancy prediction also incorporates information from other inputs, such as depth, semantic segmentation, and instance segmentation. Depth estimation is a fundamental task in camera-based autonomous driving and can be categorized into relative depth \cite{yang2024depth, godard2017unsupervised} and absolute depth \cite{huang2024selfocc, fang2024gsdc}. With monocular \cite{li2023learning, feng2022disentangling} or stereo \cite{bian2019unsupervised} images, absolute depth estimation provides the real distance between objects and the camera. Absolute depth information has been applied in camera-based 3D semantic occupancy prediction to resolve projection ambiguity in the frontal direction \cite{li2023voxformer,wang2024h2gformer}, thereby helping to determine occupancy state. 

Semantic segmentation \cite{long2015fully} predicts a semantic label for each image pixel, while instance segmentation \cite{li2017fully} distinguishes objects within a class based on it. Cityscapes \cite{cordts2016cityscapes}, KITTI \cite{abu2018augmented}, and nuScenes \cite{caesar2020nuscenes} provide diverse driving scenes and conditions. By training on these datasets, many classic models, such as FCNs \cite{long2015fully}, and DeepLab \cite{chen2017deeplab}, achieve robust and efficient segmentation for real-time application. Recently, universal segmentation models \cite{lambert2020mseg,liang2018dynamic} aim to improve scalability by training on mixed datasets.

Inspired by the above, DSOcc explores the use of depth and image semantic segmentation information to enhance camera-based 3D semantic occupancy prediction.

\section{Methodology}

\begin{figure*}[t] 
\begin{center}
\includegraphics[width=\linewidth]{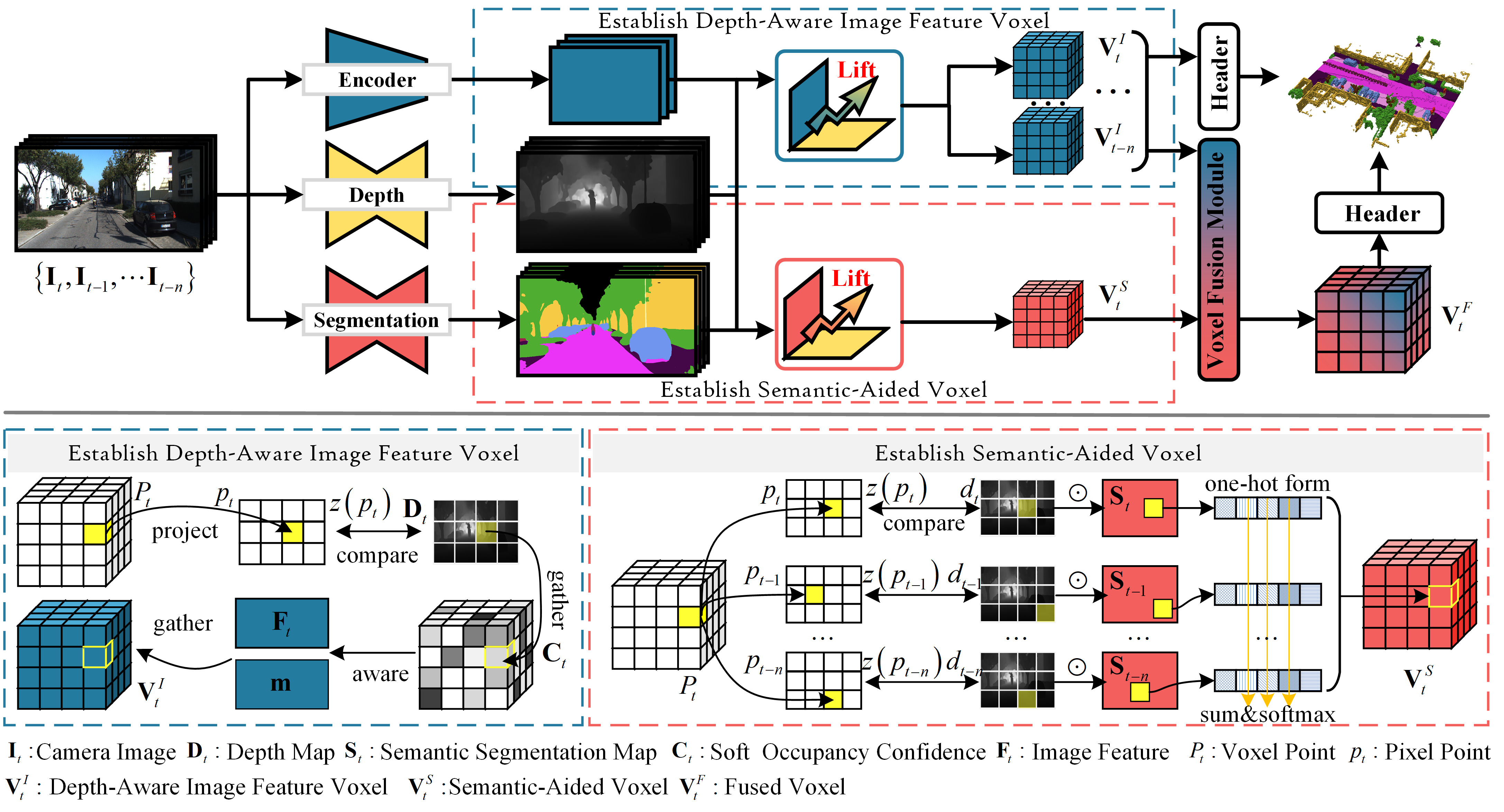}  
\end{center}
\caption{The framework of DSOcc. To address the misclassification issue in occupancy state inference, we compute a soft occupancy confidence to implicitly represent the occupancy state and embed it into image features. To overcome the limitations of occupancy class reasoning, we fuse multi-frame semantic segmentation maps for better inference capacity and class distribution. Finally, a voxel fusion module is incorporated to integrate occupancy state and class predictions for the final output.}
\label{fig2}
\end{figure*}

\subsection{Overview}
\label{sec3.1}
The framework of DSOcc is illustrated in Fig. \ref{fig2}. We aim to take the multi-frame camera images $\left\{ {{{\bf{I}}_t},{{\bf{I}}_{t - 1}}, \cdots ,{{\bf{I}}_{t - n}}} \right\} \in {{ \mathbb{R} }^{h \times w}}$ and camera parameters $\left\{ {{\bf{K}},{\bf{T}}} \right\}$ as inputs to predict the 3D semantic occupancy ${{\bf{Y}}_t} \in {\left\{ {{c_0},{c_1}, \ldots {c_M}} \right\}^{H \times W \times Z}}$ at the timestep $t$. Here, $n + 1$ is the frame number of camera images; $h,w$ denotes the height and width of camera image; ${\bf{K}},{\bf{T}}$ denote the intrinsic and extrinsic matrix of camera; ${\bf{T}} = \left[ {{\bf{R}}|{\bf{t}}} \right]$ is composed of rotation matrix ${\bf{R}}$ and translation matrix ${\bf{t}}$; $H,W,Z$ are the length, width, height of voxel grid; $M$ denotes the class number of 3D semantic occupancy. In a word, the mathematical formulation is ${{\bf{Y}}_t} = \Theta \left( {\left\{ {{{\bf{{\rm I}}}_{t - i}}} \right\}_{t = 0}^n,{\bf{{K}}},{\bf{T}}} \right)$, where $\Theta $ is the proposed DSOcc.

DSOcc aims to enhance the inference capacity of the occupancy state (whether) and the occupancy class (which) in 3D semantic occupancy prediction. It utilizes a pre-trained image encoder to extract image features ${\left\{ {{{\bf{F}}_{t - i}}} \right\}_{i = 0}^n}$, and employs a well-trained depth predictor and semantic segmentation model to infer depth maps ${\left\{ {{{\bf{D}}_{t - i}}} \right\}_{i = 0}^n}$ and semantic segmentation map ${\left\{ {{{\bf{S}}_{t - i}}} \right\}_{i = 0}^n}$. To address the challenges of ambiguous spatial correspondences between voxels and images in occupancy state inference, we lift 2D image features into depth-aware image feature voxels for each frame and obtain ${\left\{ {{\bf{V}}_{t - i}^I} \right\}_{i = 0}^n}$ by computing a soft occupancy confidence and backward project image features into a dense voxel. To aid the occupancy class inference, we lift semantic segmentation maps to a semantic-aided voxel ${{\bf{V}}_t^S}$ by integrating multi-frame one-hot representations with occupancy probabilities. Concurrently, we propose a novel voxel fusion module that fuses ${\left\{ {{\bf{V}}_{t - i}^I} \right\}_{i = 0}^n}$ and ${{\bf{V}}_t^S}$ to obtain ${{\bf{V}}_t^F}$ by utilizing global interaction through deformable attention.

\textbf{Depth Predictor}. It estimates the absolute depth between objects and ego vehicle, conditioned on camera images and calibrations. Specifically, we leverage the well-trained MobileStereoNet \cite{shamsafar2022mobilestereonet}, similar to previous method \cite{jiang2024symphonize,li2023voxformer}. This model is lightweight and well-suited for real-time applications.

\textbf{Semantic Segmentation Model}. It is pretrained on large-scale datasets, and exact class alignment with our task is not required. Specifically, we employ MSeg \cite{lambert2020mseg} and adapt its universal class classification to the specific class presets required for this task.

\subsection{Depth-Aware Image Feature Voxel}
\label{sec3.2}

\begin{figure}[t] 
\begin{center}
\includegraphics[width=\linewidth]{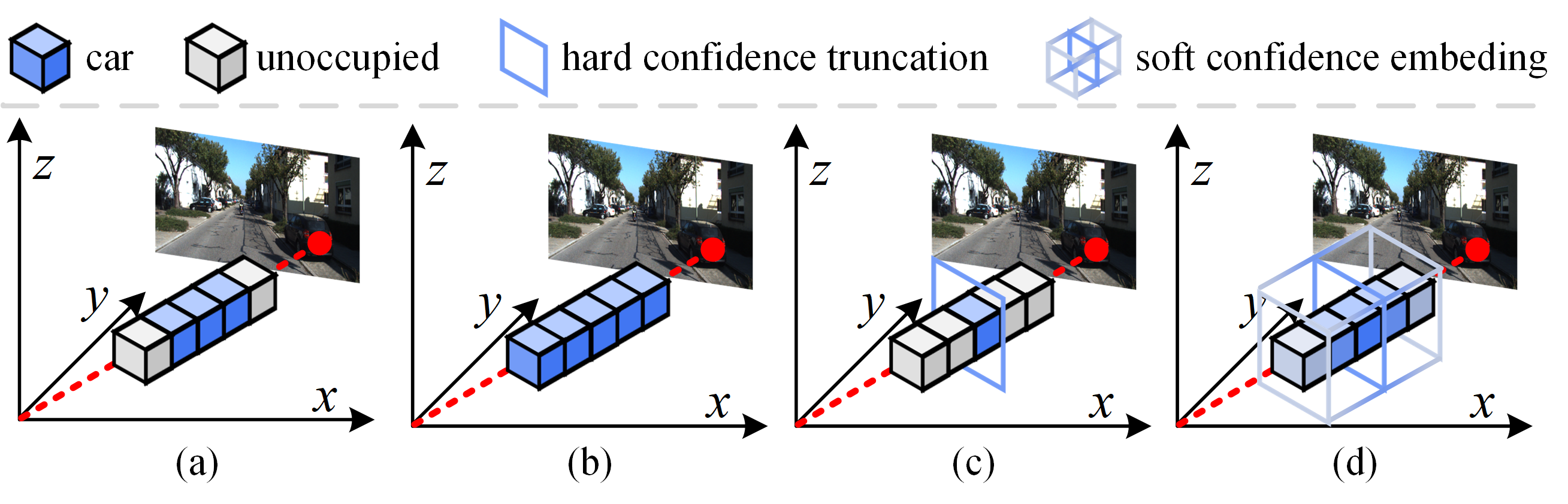}  
\end{center}
\caption{Occupancy state inference of different methods. (a) ground truth. (b) without depth information. (c) explicit occupancy state inference using hard confidence truncation. (d) our depth-aware approach: embedding a soft truncation.}
\label{fig3}
\end{figure}

The voxels are projected into the image coordinates using camera calibration parameters, and the corresponding image features are collected. Similar to previous methods \cite{cao2022monoscene,mei2024camera}, as illustrated in Fig. \ref{fig3}b, unoccupied voxels share the same projection position as neighboring occupied voxels in the absence of depth information. Consequently, these unoccupied voxels erroneously collect object-related information, leading to numerous incorrect occupancy state inferences. For previous methods \cite{li2023voxformer,wang2024h2gformer}, they learn explicit occupancy states using depth information in a separate stage. Intrinsically, as illustrated in Fig. \ref{fig3}c, these methods employ a hard confidence truncation to provide a binary estimation for each voxel. According to our calculations, these methods misclassify nearly 40\% of occupied voxels as unoccupied, resulting in these false unoccupied voxels failing to collect any features in the subsequent stages. 

Therefore, in this section, we aim to learn a soft occupancy confidence conditioned on depth information and embed it into image features. This enables the model to adaptively infer occupancy states, identifying potential unoccupied voxels while allowing all voxels to learn image features from their projected positions.

As illustrated in Fig. \ref{fig2}, let $P$ be the voxel center in the ego vehicle coordinate, and $p$ is the corresponding projected point in the image coordinate. The projection process is described as follow:

\begin{equation}
\label{equ1}
p = {\mathbf{K}}\left( \mathbf{R}P + \mathbf{t} \right)
\end{equation}

For the projected point $p=(x,y,z)$, its $x,y$ values determine the pixel position, while its $z$ value represents the distance between $P$ and the camera. We compare ${z}$ with its depth ${d}$ from ${{\bf{D}}}$. Intuitively, when the real object is located in this voxel, ${z}$ will be equal to ${d}$. Therefore, we calculate a soft occupancy confidence $c$ to represent the distance between the voxel and the real object. The mathematical expression is given by:
\begin{equation}
\label{equ2}
{c = \exp \left( { - \left| {z - d} \right|} \right)}
\end{equation}
The range of $c$ is ${\left( {0,\left. 1 \right]} \right.}$, and $c$ will approach 1 when $P$ is close to the real object at $p$, and vice versa.

Next, we aim embed this soft occupancy confidence into image features to form a depth-aware image feature voxel. For feature voxel establishment, previous methods \cite{li2023voxformer,jiang2024symphonize} use a learning-based approach to collect image features into the voxel. This requires additional computation and is prone to misalignment. In contrast, we explicitly establish the image feature voxel based on the projection relationship. We define a separate voxel for each frame and only illustrate the case at timestep $t$ for convenience. For the soft occupancy confidence, we set $c = 0$ when the projected point $p$ is outside the image dimension, and assign $c$ at the corresponding position of $P$ in voxel to obtain a depth-aware matrix ${{\bf{C}}}$. For the image features, we devise a learnable matrix ${{\bf{m}}}$ as a marker for unoccupied voxels, and our feature voxel ${{\bf{\tilde F}}}$ is described as follow:

\begin{equation}
\label{equ3}
{{\bf{\tilde F}} = \left\{ {\begin{array}{*{20}{c}}
{{{\bf{F}}^ \uparrow },p \in \left[ {0,w} \right) \times \left[ {0,h} \right)}\\
{{{\bf{m}}^ \uparrow },p \notin \left[ {0,w} \right) \times \left[ {0,h} \right)}
\end{array}} \right.}
\end{equation}

\noindent where ${^ \uparrow }$ denotes collection operation. Lastly, we embed depth-aware matrix ${{\bf{C}}}$ into feature voxel ${{\bf{\tilde F}}}$ to obtain depth-aware image feature voxel ${{{\bf{V}}^I}}$:

\begin{equation}
\label{equ4}
{{{\bf{V}}^I} = {\bf{C}} \cdot {\bf{\tilde F}}}
\end{equation}

With this depth-aware embedding, when the voxel is far from the real object, the image feature assigned to it is attenuated by the soft occupancy confidence. As the distance increases, the image feature approaches zero, allowing the model to identify it as unoccupied.

\subsection{Semantic-Aided Voxel}
\label{sec3.3}

\begin{figure*}[t] 
\begin{center}
\includegraphics[width=\linewidth]{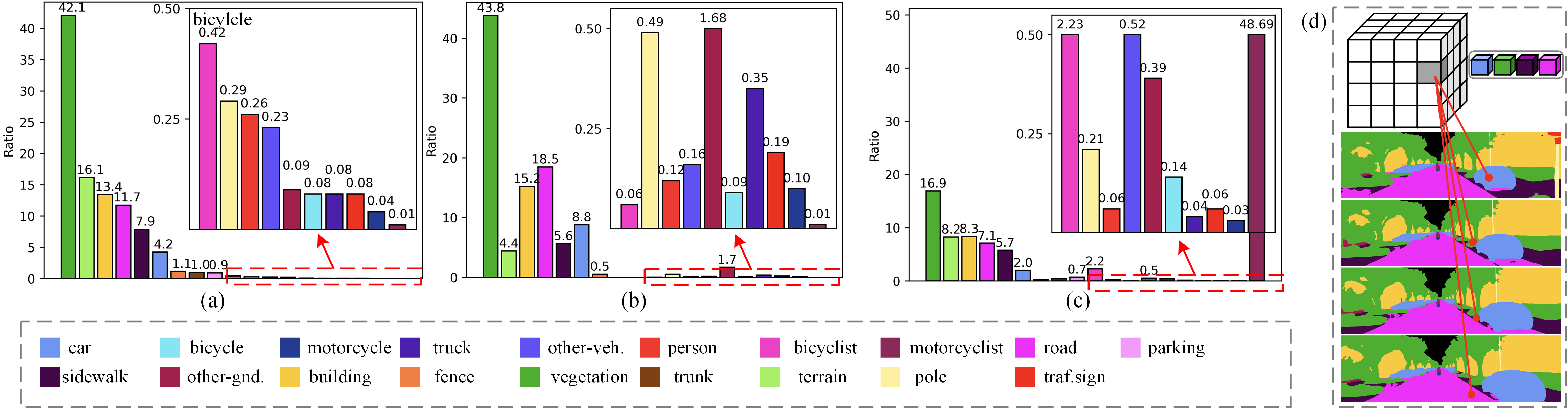}  
\end{center}
\caption{The class distribution on SemanticKITTI validation dataset \cite{behley2019semantickitti}. (a) the ground truth. (b) camera image semantic segmentation. (c) the prediction case when only use occupancy annotation during training phase. (d) projection shift between adjacent frames, A voxel may be projected onto different classes in multi-frame semantic segmentation maps. }
\label{fig4}
\end{figure*}

The core challenge in camera-based 3D semantic occupancy prediction is to identify the class of each voxel based on image features. However, due to limitations in model capacity and the dataset scale, prediction models face challenges such as class imbalance and poor class discrimination, as illustrated in Fig. \ref{fig4}c. In contrast, image semantic segmentation benefit from larger and more diverse datasets, achieving greater robustness. Fundamentally, both image semantic segmentation and 3D semantic occupancy prediction share the same objective, and the former class distribution closely resembles the latter ground truth, as illustrated in Fig. \ref{fig4}a and Fig. \ref{fig4}b. Therefore, we aim to use the former capacity in the latter task to aid occupancy class inference.

We establish a semantic-aided voxel to collect segmentation information from ${\left\{ {{{\bf{S}}_{t - i}}} \right\}_{i = 0}^n}$ using the projection relationship defined as Equ. \ref{equ1}. As illustrated in Fig. \ref{fig4}d, the same voxel may collect different classes from different frames due to the projection shift between adjacent frames. Since the results are discrete, effectively utilizing segmentation information from multiple frames is a key challenge.

To address this, we aim to fuse segmentation with their occupancy probabilities for sorting and weighted concatenation, enabling the subsequent module to adaptively extract useful segmentation despite projection shifts. Specifically, we transform ${\left\{ {{{\bf{S}}_{t - i}}} \right\}_{i = 0}^n}$ into the one-hot form, and utilize the soft occupancy confidence defined as Equ. \ref{equ2} to determine the occupancy probabilities across frames. We then multiply the occupancy probabilities by one-hot form segmentation for each frame and sum results across all frames. In this way, when the distance between the voxel and the real object in one frame is smaller, the occupancy class of this voxel should more closely align with the segmentation of this frame. As illustrated in Fig. \ref{fig2}, we obtain the semantic-aided voxel ${{\bf{V}}_t^S}$ at the timestep ${t}$ as:

\begin{equation}
\label{equ5}
{{\bf{V}}_t^S = {\rm{softmax}}\left[ {\sum\limits_{i = 0}^n {\left( {{{\bf{C}}_{t - i}} \cdot {{\bf{S}}_{t - i}}^ \uparrow } \right)} } \right]}
\end{equation}

\subsection{Voxel Fusion Module}
\label{sec3.4}

In this section, we aim to fuse depth-aware image feature voxels ${\left\{ {{\bf{V}}_{t - i}^I} \right\}_{i = 0}^n}$and semantic-aided voxel ${{\bf{V}}_t^S}$ to obtain a fused voxel ${{\bf{V}}_t^F}$ for final prediction. As illustrated in Fig. \ref{fig4}, initially, ${\left\{ {{\bf{V}}_{t - i}^I} \right\}_{i = 0}^n}$ are concatenated along the channel dimension and projected into ${{\bf{\tilde V}}_t^I}$ through a \texttt{Conv3D} layer, which maintains the original voxel size but reduces the channel number from ${nC}$ to ${C}$. Similarly, ${{\bf{V}}_t^S}$ is projected into ${{\bf{\tilde V}}_t^S}$, where the channel number is increased from ${M}$ to ${C}$. Although the depth-aware image feature voxel and the semantic-aided voxel are aligned for each voxel, the model still needs to learn global spatial cues for prediction.

\begin{figure}[t] 
\begin{center}
\includegraphics[width=\linewidth]{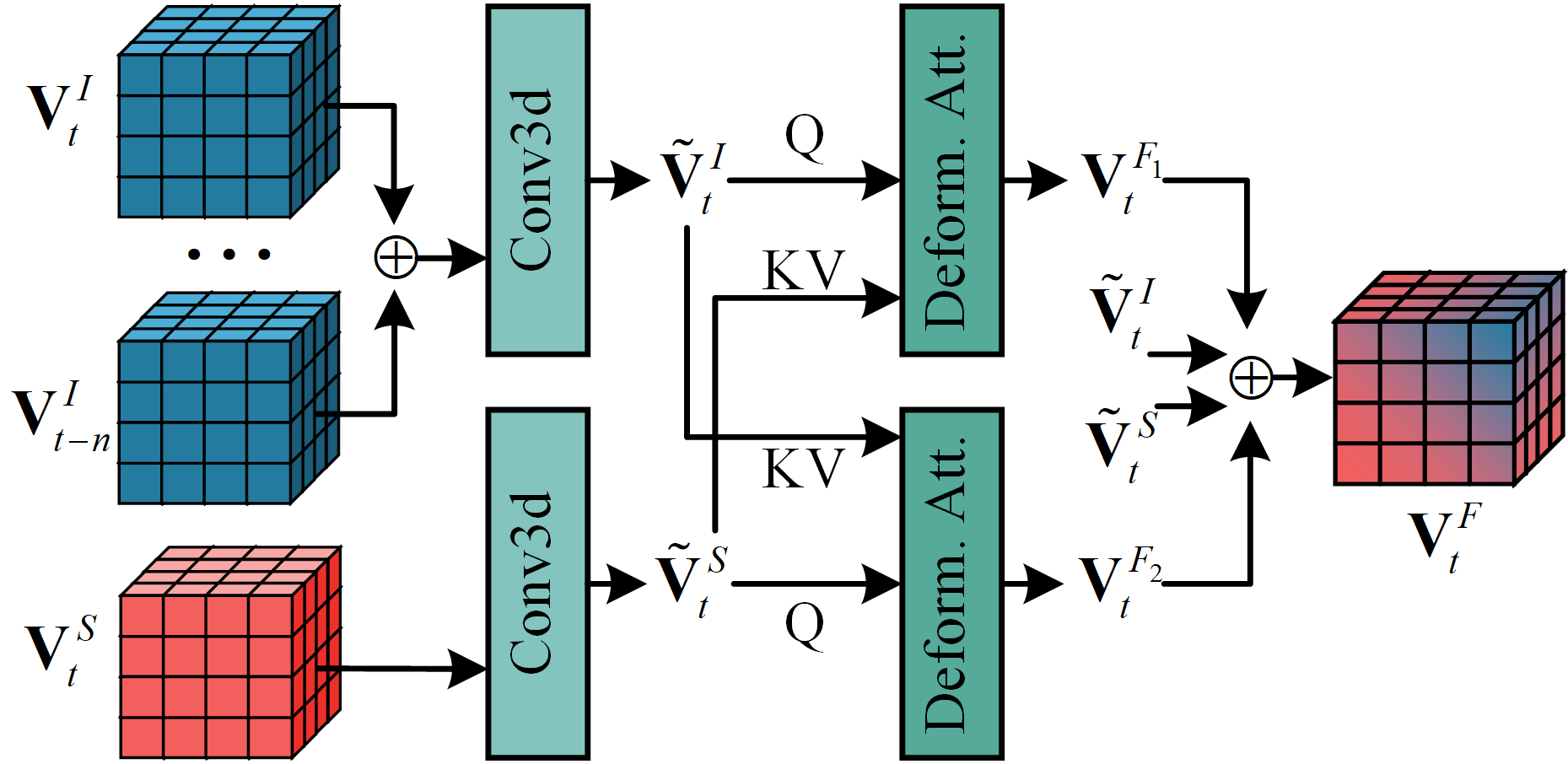}  
\end{center}
\caption{Voxel fusion module: The dual interaction mechanism aggregates segmentation and image feature information globally via deformable attention to obtain the fused voxel. }
\label{fig5}
\end{figure}

Since voxel interaction has cubic complexity, we employ the computation-efficient deformable attention (\texttt{DA}) \cite{zhudeformable} to learn global spatial cues. \texttt{DA} aggregates the feature ${{\bf{F}}}$ into the query ${{\bf{q}}}$ with several deformable reference points ${p}$. The mathematical expression is given by: 

\begin{equation}
\label{equ6}
{\texttt{DA}\left( {{\bf{q}},p,{\bf{F}}} \right) = \sum\limits_{s = 1}^{{N_s}} {{{\bf{A}}_s}{{\bf{W}}_s}{\bf{F}}\left( {p + \Delta {p_s}} \right)} }
\end{equation}

\noindent where ${{N_s}}$ denotes the number of sampling points, ${{{\bf{A}}_s} \in \left[ {0,1} \right]}$ denotes a learnable weight matrix conditioned on ${{\bf{q}}}$ for the sampling point ${s}$, ${{{\bf{W}}_s} \in {{ \mathbb{R} }^{C \times C}}}$ denotes the projection matrix for value transformation, and ${\Delta {p_s}}$ is the predicted offset relative to ${p}$ conditioned on ${{\bf{q}}}$.

Specifically, we propose a dual interaction mechanism for ${{\bf{\tilde V}}_t^S}$ and ${{\bf{\tilde V}}_t^I}$. ${{\bf{\tilde V}}_t^I}$ is utilized as a query to aggregate segmentation information from ${{\bf{\tilde V}}_t^S}$ with the voxel reference point ${P}$ to obtain ${{\bf{V}}_t^{{F_1}}}$. Conversely, ${{\bf{\tilde V}}_t^S}$ is utilized as query to aggregate image feature information from ${{\bf{\tilde V}}_t^I}$ to obtain ${{\bf{V}}_t^{{F_2}}}$. This interaction process is described as: 

\begin{equation}
\label{equ7}
{\left\{ \begin{array}{l}
{\bf{V}}_t^{{F_1}} = \texttt{DA}\left( {{\bf{\tilde V}}_t^I,P,{\bf{\tilde V}}_t^S} \right)\\
{\bf{V}}_t^{{F_2}} = \texttt{DA}\left( {{\bf{\tilde V}}_t^S,P,{\bf{\tilde V}}_t^I} \right)
\end{array} \right.}
\end{equation}

At the end of voxel fusion module, we concatenate ${{\bf{\tilde V}}_t^I}$,${{\bf{V}}_t^{{F_1}}}$,${{\bf{V}}_t^{{F_2}}}$,${{\bf{\tilde V}}_t^S}$ along the channel dimension to obtain the fused voxel ${{\bf{V}}_t^F}$.

\subsection{Losses}
\label{sec3.5}

During the training phase, we use two headers consisting of \texttt{Conv3D} layers to predict ${{\bf{Y}}_t}$ and ${{{\bf{Y}}_{ti}}}$ conditioned on ${{\bf{V}}_t^F}$ and ${{\bf{\tilde V}}_t^I}$ at the timestep ${t}$, respectively. The prediction of ${{{\bf{Y}}_{ti}}}$ supervises the extraction and learning of image features. Similar to previous methods \cite{jiang2024symphonize,li2023voxformer,mei2024camera}, we employ the standard semantic loss ${{{\cal L}_{sem}}}$ and geometric loss ${{{\cal L}_{geo}}}$ as defined in Monoscene \cite{cao2022monoscene} to supervise predictions $\left( {{{\bf{Y}}_t},{{\bf{Y}}_{ti}}} \right)$ at the semantic and geometric level, respectively. We also use the weighted cross-entropy loss ${{{\cal L}_{ce}}}$, where the weight is calculated from class frequency. The overall loss function is defined as:

\begin{equation}
\label{equ8}
\mathcal{L} = \sum_{x \in \{t, ti\}} \big( \mathcal{L}_{\text{sem}}(\mathbf{Y}_x, \hat{\mathbf{Y}}_t) 
+ \mathcal{L}_{\text{geo}}(\mathbf{Y}_x, \hat{\mathbf{Y}}_t) + \mathcal{L}_{\text{ce}}(\mathbf{Y}_x, \hat{\mathbf{Y}}_t) \big)
\end{equation}

\noindent where ${{\bf{\hat Y}}_t}$ denotes the ground truth for camera-based 3D semantic occupancy prediction. Additionally, there are no hyperparameters to balance the different losses and prediction branches, ensuring its scalability.

During the testing phase, we select ${{\bf{Y}}_t}$ as the final prediction, and the header for predicting ${{{\bf{Y}}_{ti}}}$ is no longer needed.

\section{Experiment}
\subsection{Dataset}
\label{sec4.1}
\textbf{Dataset}. We evaluate DSOcc on the SemanticKITTI \cite{behley2019semantickitti}, SSCBenchKITTI-360 \cite{li2024sscbench}, and Occ3D-nuScenes \cite{tian2023occ3d} datasets. SemanticKITTI and SSCBenchKITTI-360 focus on a scene of ${51.2m \times 51.2m \times 6.4m}$, with a voxel size of ${0.2m \times 0.2m \times 0.2m}$. SemanticKITTI includes 20 class labels (19 semantics + 1 free) , whereas SSCBenchKITTI-360 includes 19 class labels (18 semantics + 1 free). Occ3D-nuScenes dataset comprises 700 training scenes and 150 validation scenes, each annotated within an ${80m \times 80m \times 6.4m}$ volume at a voxel resolution ${0.4m \times 0.4m \times 0.4m}$ for 18 semantic classes.

\textbf{Metric}. Following SSC, we measure performance using intersection over union (IoU) and mean IoU (mIoU) over 19 classes. IoU reflects overall scene completion quality, while mIoU the quality of semantic segmentation for each class. We report mIoU as our primary metric.

\subsection{Implementation Details}
\label{sec4.2}
We crop images to sizes ${1220 \times 370}$, ${1408 \times 376}$, and ${1600 \times 900}$ for the SemanticKITTI, SSCBenchKITTI-360, and Occ3D-nuScenes datasets, respectively, following previous methods \cite{li2023voxformer, mei2024camera, zhang2023occformer,li2024bevformer}. We use ResNet50 \cite{he2016deep} to extract its feature map. All experiments were conducted on 2 NVIDIA A6000 GPUs with a batch size of 1. The sizes of depth-aware image feature voxel and semantic-aided voxel are set to ${{1 \mathord{\left/{\vphantom {1 2}} \right.\kern-\nulldelimiterspace} 2}}$ as the final prediction. The \texttt{DA} used in the voxel fusion module has 1 layer and 8 sampling points. We optimize DSOcc using the AdamW optimizer \cite{loshchilov2017decoupled} with an initial learning rate of ${2 \times {e^{ - 4}}}$ and a weight decay of ${1 \times {e^{ - 4}}}$ for a total of 40 epochs.  Following the method in \cite{li2023voxformer,wang2024not,mei2024camera,wang2024h2gformer}, we propose two versions of the DSOcc model on SemanticKITTI and SSCBenchKITTI-360 datasets: \textbf{DSOcc-T}, which takes the current and previous four frames as inputs, and \textbf{DSOcc-S}, which takes only the current frame as input.

\begin{table*}[!htb]
\centering
\setlength{\tabcolsep}{1mm} 
\caption{\textbf{The quantitative results on the SemanticKITTI validation set. \textbf{Bold}} and \underline{underline} represent the best and second best results, respectively. Pub. indicates the year of publication and the journal or conference in which the work appeared.}
\resizebox{\linewidth}{!}{%
\begin{tabular}{l|l|l>{\columncolor{gray!30}}l|lllllllllllllllllll}
\toprule
Method & Pub. & IoU & mIoU &\rotatebox{90}{{\tikz \fill[c1] (0,0) rectangle (0.6em,0.6em);} car (3.92\%)} & \rotatebox{90}{{\tikz \fill[c2] (0,0) rectangle (0.6em,0.6em);} bicycle (0.03\%)} & \rotatebox{90}{{\tikz \fill[c3] (0,0) rectangle (0.6em,0.6em);} motorcycle (0.03\%)} & \rotatebox{90}{{\tikz \fill[c4] (0,0) rectangle (0.6em,0.6em);} truck (0.16\%)} & \rotatebox{90}{{\tikz \fill[c5] (0,0) rectangle (0.6em,0.6em);} other-veh. (0.20\%)} & \rotatebox{90}{{\tikz \fill[c6] (0,0) rectangle (0.6em,0.6em);} person (0.07\%)} & \rotatebox{90}{{\tikz \fill[c7] (0,0) rectangle (0.6em,0.6em);} bicyclist (0.07\%)} & \rotatebox{90}{{\tikz \fill[c8] (0,0) rectangle (0.6em,0.6em);} motorcyclist (0.05\%)} & \rotatebox{90}{{\tikz \fill[c9] (0,0) rectangle (0.6em,0.6em);} road (15.30\%)} & \rotatebox{90}{{\tikz \fill[c10] (0,0) rectangle (0.6em,0.6em);} parking (1.12\%)} & \rotatebox{90}{{\tikz \fill[c11] (0,0) rectangle (0.6em,0.6em);} sidewalk (11.13\%)} & \rotatebox{90}{{\tikz \fill[c12] (0,0) rectangle (0.6em,0.6em);} other-grnd.(0.56\%)} & \rotatebox{90}{{\tikz \fill[c13] (0,0) rectangle (0.6em,0.6em);} building (14.10\%)} & \rotatebox{90}{{\tikz \fill[c14] (0,0) rectangle (0.6em,0.6em);} fence (3.90\%)} & \rotatebox{90}{{\tikz \fill[c15] (0,0) rectangle (0.6em,0.6em);} vegetation (39.3\%)} & \rotatebox{90}{{\tikz \fill[c16] (0,0) rectangle (0.6em,0.6em);} trunk (0.51\%)} & \rotatebox{90}{{\tikz \fill[c17] (0,0) rectangle (0.6em,0.6em);} terrain (9.17\%)} & \rotatebox{90}{{\tikz \fill[c18] (0,0) rectangle (0.6em,0.6em);} pole (0.29\%)} & \rotatebox{90}{{\tikz \fill[c19] (0,0) rectangle (0.6em,0.6em);} traf.-sign (0.08\%)} \\ 
\midrule
TPVFormer \cite{huang2023tri} & CVPR23 & 35.61 & 11.36 & 23.81 & 0.36 & 0.05 & 8.08 & 4.35 & 0.51 & 0.89 & 0.00 & 56.50 & 20.60 & 25.87 & \underline{0.85} & 13.88 & 5.94 & 16.92 & 2.26 & 30.38 & 3.14 & 1.52 \\
OccFormer \cite{zhang2023occformer}& CVPR23 & 36.50 & 13.46 & 25.09 & 0.81 & 1.19 & \underline{25.53} & 8.52 & 2.78 & \underline{2.82} & 0.00 & 58.85 & 19.61 & 26.88 & 0.31 & 14.40 & 5.61 & 19.63 & 3.93 & 32.62 & 4.26 & 2.86 \\
VoxFormer-T \cite{li2023voxformer} & CVPR23 & 44.15 & 13.35 & 25.64 & 1.28 & 0.56 & 7.26 & 7.81 & 1.93 & 1.97 & 0.00 & 53.57 & 19.69 & 26.52 & 0.42 & 19.54 & 7.31 & 26.10 & 6.10 & 33.06 & 9.15 & 4.94 \\
PanoSSC \cite{shi2024panossc}& 3DV24 & 34.94 & 11.22 & 19.63 & 0.63 & 0.36 & 14.79 & 6.22 & 0.87 & 0.00 & 0.00 & 56.36 & 17.76 & 26.40 & \textbf{0.88} & 14.26 & 5.72 & 16.69 & 1.83 & 28.05 & 1.94 & 0.70 \\
H2GFormer-T \cite{wang2024h2gformer}& AAAI24 & 44.69 & 14.29 & 28.21 & 0.95 & 0.91 & 6.80 & 9.32 & 1.15 & 0.10 & 0.00 & 57.00 & \underline{21.74} & 29.37 & 0.34 & 20.51 & 7.98 & 27.44 & 7.80 & 36.26 & 9.88 & 5.81 \\
OctreeOcc \cite{lu2023octreeocc}& Arix24 & \underline{44.71} & 13.12 & 28.07 & 0.64 & 0.71 & 16.43 & 6.03 & 2.25 & 2.57 & 0.00 & 55.13 & 18.68 & 26.74 & 0.65 & 18.69 & 4.01 & 25.26 & 4.89 & 32.47 & 3.72 & 2.36 \\
SGN-T \cite{mei2024camera}& TIP24 & \textbf{46.21} & 15.32 & \underline{33.31} & 0.61 & 0.46 & 6.03 & \underline{9.84} & 0.47 & 0.10 & 0.00 & 59.10 & 19.05 & 29.41 & 0.33 & \underline{25.17} & \underline{9.96} & \textbf{28.93} & \underline{9.58} & \textbf{38.12} & \textbf{13.25} & \underline{7.32} \\
Symphonies \cite{jiang2024symphonize}& CVPR24 & 41.44 & 13.44 & 27.23 & \underline{1.44} & \underline{2.28} & 15.99 & 9.52 & \underline{3.19} & \textbf{8.09} & 0.00 & 55.78 & 14.57 & 26.77 & 0.19 & 18.76 & 6.18 & 24.50 & 4.32 & 28.49 & 8.99 & 5.39 \\
HASSC-T \cite{wang2024not}& CVPR24 & 44.55 & \underline{15.88} & 30.64 & 1.20 & 0.91 & 23.72 & 7.77 & 1.79 & 2.47 & 0.00 & \textbf{62.75} & 20.20 & \underline{32.40} & 0.51 & 22.90 & 8.67 & 26.47 & 7.14 & \underline{38.10} & 9.00 & 5.23 \\  
\midrule
DSOcc-T(Ours) &  & 44.41 & \textbf{18.02} & \textbf{34.08} & \textbf{1.45} & \textbf{7.63} & \textbf{26.60} & \textbf{13.86} & \textbf{5.25} & 0.56 & 0.00 & \underline{60.82} & \textbf{21.76} & \textbf{33.08} & 0.46 & \textbf{25.30} & \textbf{11.59} & \underline{28.73} & \textbf{12.21} & 37.52 & \underline{12.50} & \textbf{8.87} \\ 
\bottomrule
VoxFormer-S \cite{li2023voxformer} & CVPR23 & 44.04 & 12.35 &25.79 & 0.59 & 0.51 &5.63 &3.77 &1.78 &\underline{3.32} & 0.00 &54.76 &15.50 &26.35 & \underline{0.70} &17.65 & 7.64 & 24.39 &5.08 &29.96 &7.11 &4.18 \\

H2GFormer-S \cite{wang2024h2gformer}& AAAI24 & \underline{44.57} & 13.73 & 27.60 &0.50 &0.47 &\underline{10.00} &\textbf{7.39} &1.54 &2.88 &0.00 &56.08 &17.83 &29.12 &0.45 &19.74 & 7.24 &\textbf{26.25} &\underline{6.80} &\textbf{34.42} &7.88 &4.68 \\

HASSC-S \cite{wang2024not}& CVPR24 & \textbf{44.82} &13.48 &27.23 &\textbf{0.92} &\underline{0.86} &9.91 &5.61 &\textbf{2.80} &\textbf{4.71} &0.00 &57.05 &15.90 &28.25 &\textbf{1.04} &19.05 &6.58 &25.48 &6.15 &\underline{32.94} &7.68 &4.05 \\  

SGN-S \cite{mei2024camera}& TIP24 & 43.60 & \underline{14.55} & \underline{31.88} &0.58 &0.17 &\textbf{13.18} &5.68 &1.28 &1.49 &0.00 &\underline{59.32} &\textbf{18.46} &\underline{30.51} &0.42 &\textbf{21.43} &\underline{9.66} &\underline{25.98} &7.43 &\textbf{34.42} &\textbf{9.83} &\underline{4.71} \\
\midrule
DSOcc-S(Ours) &  &42.67 &\textbf{14.59} &\textbf{31.89} &\underline{0.68} &\textbf{3.67} &7.30 &\underline{6.06} &\underline{2.40} &0.86 & 0.00 &\textbf{60.30} &\underline{18.42} &\textbf{31.92} &0.53 &\underline{20.37} &\textbf{11.85} &25.53  &\textbf{7.81} &32.52 &\underline{9.68} &\textbf{5.46} \\ 
\bottomrule
\end{tabular}
}

\label{table1}
\end{table*}

\begin{table*}[!htb]
\centering
\caption{\textbf{The Quantitative results on the SemanticKITTI validation set}. \textbf{Bold} and \underline{underline} represent the best and second best results, respectively. Pub. indicates the year of publication and the journal or conference in which the work appeared.}
\setlength{\tabcolsep}{1mm} 
\resizebox{\linewidth}{!}{%
\begin{tabular}{l|l|l>{\columncolor{gray!30}}l|lllllllllllllllllll}
\toprule
Method & Pub. & IoU & mIoU &\rotatebox{90}{{\tikz \fill[c1] (0,0) rectangle (0.6em,0.6em);} car (3.92\%)} & \rotatebox{90}{{\tikz \fill[c2] (0,0) rectangle (0.6em,0.6em);} bicycle (0.03\%)} & \rotatebox{90}{{\tikz \fill[c3] (0,0) rectangle (0.6em,0.6em);} motorcycle (0.03\%)} & \rotatebox{90}{{\tikz \fill[c4] (0,0) rectangle (0.6em,0.6em);} truck (0.16\%)} & \rotatebox{90}{{\tikz \fill[c5] (0,0) rectangle (0.6em,0.6em);} other-veh. (0.20\%)} & \rotatebox{90}{{\tikz \fill[c6] (0,0) rectangle (0.6em,0.6em);} person (0.07\%)} & \rotatebox{90}{{\tikz \fill[c7] (0,0) rectangle (0.6em,0.6em);} bicyclist (0.07\%)} & \rotatebox{90}{{\tikz \fill[c8] (0,0) rectangle (0.6em,0.6em);} motorcyclist (0.05\%)} & \rotatebox{90}{{\tikz \fill[c9] (0,0) rectangle (0.6em,0.6em);} road (15.30\%)} & \rotatebox{90}{{\tikz \fill[c10] (0,0) rectangle (0.6em,0.6em);} parking (1.12\%)} & \rotatebox{90}{{\tikz \fill[c11] (0,0) rectangle (0.6em,0.6em);} sidewalk (11.13\%)} & \rotatebox{90}{{\tikz \fill[c12] (0,0) rectangle (0.6em,0.6em);} other-grnd.(0.56\%)} & \rotatebox{90}{{\tikz \fill[c13] (0,0) rectangle (0.6em,0.6em);} building (14.10\%)} & \rotatebox{90}{{\tikz \fill[c14] (0,0) rectangle (0.6em,0.6em);} fence (3.90\%)} & \rotatebox{90}{{\tikz \fill[c15] (0,0) rectangle (0.6em,0.6em);} vegetation (39.3\%)} & \rotatebox{90}{{\tikz \fill[c16] (0,0) rectangle (0.6em,0.6em);} trunk (0.51\%)} & \rotatebox{90}{{\tikz \fill[c17] (0,0) rectangle (0.6em,0.6em);} terrain (9.17\%)} & \rotatebox{90}{{\tikz \fill[c18] (0,0) rectangle (0.6em,0.6em);} pole (0.29\%)} & \rotatebox{90}{{\tikz \fill[c19] (0,0) rectangle (0.6em,0.6em);} traf.-sign (0.08\%)} \\ 
\midrule
TPVFormer \cite{huang2023tri} & CVPR23 & 34.25 & 11.26 & 19.20 & 1.00 & 0.50 & 3.70 & 2.30 & 1.10 & 2.40 & 0.30 & 55.10 & 27.40 & 27.20 & 6.50 & 14.80 & 11.00 & 13.90 & 2.60 & 20.40 & 2.90 & 1.50 \\
OccFormer \cite{zhang2023occformer}  & CVPR23 & 34.53 & 12.32 & 21.30 & 1.50 & 1.70 & 3.90 & 3.20 & 2.20 & 1.10 & 0.20 & 55.90 & \underline{31.50} & 30.30 & 6.50 & 15.70 & 11.90 & 16.80 & 3.90 & 21.30 & 3.80 & 3.70 \\
VoxFormer-T \cite{li2023voxformer} & CVPR23 & 43.21 & 13.41 & 21.70 & 1.90 & 1.60 & 3.60 & 4.10 & 1.60 & 1.10 & 0.00 & 54.10 & 25.10 & 26.90 & 7.30 & 23.50 & 13.10 & 24.40 & 8.10 & 24.20 & 6.60 & 5.70 \\
SurroundOcc \cite{wei2023surroundocc} & CVPR23 & 34.72 & 11.86 & 20.30 & 1.60 & 1.20 & 1.40 & 4.40 & 1.40 & 2.00 & 0.10 & 56.90 & 30.20 & 28.30 & 6.80 & 15.20 & 11.30 & 14.90 & 3.40 & 19.30 & 3.90 & 2.40 \\
H2GFormer-T \cite{wang2024h2gformer} & AAAI24 & 43.52 & 14.60 & 23.70 & 0.60 & 1.20 & \underline{5.20} & 5.00 & 1.10 & 0.10 & 0.00 & 57.90 & 30.00 & 30.40 & 6.90 & 24.00 & 14.60 & 25.20 & 10.70 & 25.80 & 7.50 & 7.10 \\
SGN-T \cite{mei2024camera}  & TIP24 & \textbf{45.42} & 15.76 & \underline{25.40} & \textbf{4.50} & 0.90 & 4.50 & 3.70 & 0.50 & 0.30 & 0.10 & \underline{60.40} & 28.90 & 31.40 & 8.70 & \textbf{28.40} & 18.10 & \underline{27.40} & \underline{12.60} & 28.40 & \underline{10.00} & 8.30 \\
LowRankOcc \cite{zhao2024lowrankocc} & CVPR24 & 38.47 & 13.56 & 20.90 & 3.30 & 2.70 & 2.90 & 4.40 & 2.40 & 1.70 & \textbf{2.30} & 52.80 & 25.10 & 27.20 & 8.80 & 22.10 & 14.40 & 22.90 & 8.90 & 20.80 & 7.00 & 7.00 \\
Symphonies \cite{jiang2024symphonize} & CVPR24 & 42.19 & 15.04 & 23.60 & \underline{3.60} & 2.60 & 3.20 & \underline{5.60} & \underline{3.20} & 1.90 & \underline{2.00} & 58.40 & 26.90 & 29.30 & \underline{11.70} & 24.70 & 16.10 & 24.20 & 10.00 & 23.10 & 7.70 & 8.00 \\
HASSC-T \cite{wang2024not} & CVPR24 & 42.87 & 14.38 & 23.00 & 1.90 & 1.50 &2.90 & 4.90 & 1.40 & \underline{3.00} & 0.00 & 55.30 & 25.90 & 29.60 & 11.30 & 23.10 & 14.30 & 24.80 & 9.80 & 26.50 & 7.00 & 7.10 \\
Bi-SSC \cite{xue2024bi} & CVPR24 & 45.10 & \underline{16.73} & 25.00 & 1.80 & \underline{2.90} & \textbf{6.80} & \textbf{6.80} & 1.70 & \textbf{3.30} & 1.00 & \textbf{63.40} & \textbf{31.70} & \textbf{33.30} & 11.20 & 26.60 & \textbf{19.40} & 26.10 & 10.50 & \underline{28.90} & 9.30 & \underline{8.40} \\ 
\midrule
DSOcc-T (Ours) & & \underline{45.12} & \textbf{16.90} & \textbf{25.60} & 2.50 &  \textbf{3.80} & 4.00 & 5.20 & \textbf{3.50} & 0.70 & 0.00 & 59.50 & 29.50 & \underline{31.80} & \textbf{13.10} & \underline{27.70} & \underline{18.20} & \textbf{31.20} & \textbf{13.90} & \textbf{30.30} & \textbf{10.60} & \textbf{9.80}\\
\bottomrule

VoxFormer-S \cite{li2023voxformer} & CVPR23 & \underline{42.95} & 12.20 & 20.80 & 1.00 & 0.70 & \underline{3.50} &3.70 & \textbf{1.40} & \textbf{2.60} & \underline{0.20} & 53.90 & 21.10 & 25.30 & \underline{5.60} & 19.80 & 11.10 & 22.40 & 7.50 & 21.30 & 5.10 & 4.90 \\
HASSC-S \cite{wang2024not} & CVPR24 &\textbf{44.20} & 13.72 & 23.40 & \underline{0.80} & \underline{0.90} & \textbf{4.80} & \textbf{4.10} & 1.20 & \underline{2.50} & 0.10 & 56.40 & \underline{26.50} & 28.60 & 4.90 & 22.80 & 13.30 & \underline{24.60} & 9.10 & 23.80 & 6.40 & \underline{6.30} \\
 SGN-S \cite{mei2024camera}  & TIP24 & 41.88 & \underline{14.01} & \textbf{24.90} & 0.40 & 0.30 &2.70 & \underline{4.00} & 1.10 & \underline{2.50} & \textbf{0.30} & \underline{57.80} & \textbf{27.70} & \underline{29.20} & 5.20 & \underline{23.90} & \underline{14.20} & 24.20 & \textbf{10.00} & \underline{25.80} & \underline{7.40} & 4.40 \\

 \midrule
 DSOcc-S (Ours) & &41.21 & \textbf{14.96} &\underline{23.70} &\textbf{1.30} &\textbf{1.30} &2.70 &3.90 &\underline{1.30} &0.90 &0.00 &\textbf{60.50} &26.20 &\textbf{29.50} &\textbf{12.10} &\textbf{24.00} &\textbf{16.60} &\textbf{26.90} &9.30 &\textbf{27.70} &\textbf{7.90} &\textbf{8.40}\\
\bottomrule
\end{tabular}
}

\label{table2}
\end{table*}

\begin{table*}[htb]
\caption{\textbf{The quantitative results on  SSCBench-KITTI-360 validation set.} \textbf{Bold} and \underline{underline} represent the best and second best results, respectively. Pub. indicates the year of publication and the journal or conference in which the work appeared.}
\centering
\setlength{\tabcolsep}{1mm} 
\resizebox{\linewidth}{!}{%
\begin{tabular}{l|l|l>{\columncolor{gray!30}}l|llllllllllllllllll}
\toprule
Method & Pub. & IoU & mIoU &\rotatebox{90}{{\tikz \fill[c1] (0,0) rectangle (0.6em,0.6em);} car (2.85\%)} & \rotatebox{90}{{\tikz \fill[c2] (0,0) rectangle (0.6em,0.6em);} bicycle (0.01\%)} & \rotatebox{90}{{\tikz \fill[c3] (0,0) rectangle (0.6em,0.6em);} motorcycle (0.01\%)} & \rotatebox{90}{{\tikz \fill[c4] (0,0) rectangle (0.6em,0.6em);} truck (0.16\%)} & \rotatebox{90}{{\tikz \fill[c5] (0,0) rectangle (0.6em,0.6em);} other-veh. (5.75\%)} & \rotatebox{90}{{\tikz \fill[c6] (0,0) rectangle (0.6em,0.6em);} person (0.02\%)} &\rotatebox{90}{{\tikz \fill[c9] (0,0) rectangle (0.6em,0.6em);} road (14.98\%)} & \rotatebox{90}{{\tikz \fill[c10] (0,0) rectangle (0.6em,0.6em);} parking (2.31\%)} & \rotatebox{90}{{\tikz \fill[c11] (0,0) rectangle (0.6em,0.6em);} sidewalk (6.43\%)} & \rotatebox{90}{{\tikz \fill[c12] (0,0) rectangle (0.6em,0.6em);} other-grnd.(2.05\%)} & \rotatebox{90}{{\tikz \fill[c13] (0,0) rectangle (0.6em,0.6em);} building (15.67\%)} & \rotatebox{90}{{\tikz \fill[c14] (0,0) rectangle (0.6em,0.6em);} fence (0.96\%)} & \rotatebox{90}{{\tikz \fill[c15] (0,0) rectangle (0.6em,0.6em);} vegetation (41.99\%)}  & \rotatebox{90}{{\tikz \fill[c17] (0,0) rectangle (0.6em,0.6em);} terrain (7.10\%)} & \rotatebox{90}{{\tikz \fill[c18] (0,0) rectangle (0.6em,0.6em);} pole (0.22\%)} & \rotatebox{90}{{\tikz \fill[c19] (0,0) rectangle (0.6em,0.6em);} traf.-sign (0.06\%)} & \rotatebox{90}{{\tikz \fill[c20] (0,0) rectangle (0.6em,0.6em);} other-struct. (4.33\%)}  & \rotatebox{90}{{\tikz \fill[c21] (0,0) rectangle (0.6em,0.6em);} other-obj. (0.28\%)}\\ 
\midrule
TPVFormer \cite{huang2023tri}         & CVPR23 & 40.22 & 13.64 & 21.56 & 1.09 & 1.37 & 8.06 & 2.57 & 2.38 & 52.99 & 11.99 & 31.07 & 3.78 & 34.83 & 4.80 & 30.08 & 17.52 & 7.46 & 5.86 & 5.48 & 2.70 \\
OccFormer \cite{zhang2023occformer}   & CVPR23 & 40.27 & 13.81 & 22.58 & 0.66 & 0.26 & 9.69 & 3.82 & 2.77 & 54.30 & 13.44 & 31.53 & 3.55 &36.42 & 4.80 & 31.00 & 19.51 & 7.77 & 8.51 & 6.95 & 4.60 \\
VoxFormer-T \cite{li2023voxformer}    & CVPR23 & 38.76 & 11.91 & 17.84 & 1.16 & 0.89 & 4.56 & 2.06 & 1.63 & 47.01 & 9.67 & 27.21 & 2.89 & 31.18 & 4.97 & 28.99 & 14.69 & 6.51 & 6.92 & 3.79 & 2.43 \\
SGN-T \cite{mei2024camera}            & TIP24  & \textbf{47.06} & \underline{18.25} & \underline{29.03} &\underline{3.43} & 2.90 & 10.89 & 5.20 & 2.99 & \textbf{58.14} & \textbf{15.04} & \textbf{36.40} & 4.43 & \textbf{42.02} & 7.72 & \underline{38.17} & \underline{23.22} & \textbf{16.73} & \textbf{16.38} & \underline{9.93} & \underline{5.86} \\
Symphonies \cite{jiang2024symphonize} & CVPR24 & 44.12 & \textbf{18.58} & \textbf{30.02} & 1.85 & \textbf{5.90} & \textbf{25.07} & \textbf{12.06} & \textbf{8.20} &54.94 & 13.83 & 32.76 & \textbf{6.93} & 35.11 &\textbf{8.58} & \textbf{38.33} & 11.52 & \underline{14.01} & 9.57 & \textbf{14.44} & \textbf{11.28} \\
GaussianFormer \cite{huang2025gaussianformer} & ECCV24 & 35.38 & 12.92 &18.93 & 1.02 &\underline{4.62} & \underline{18.07} & \underline{7.59} & 3.36 & 45.47 & 10.89 & 25.03 & \underline{5.32} & 28.44 & 5.68 & 29.54 & 8.62 & 2.99 & 2.32 & 9.51 & 5.14 \\
\midrule
DSOcc-T (Ours) & &\underline{45.12}  &17.88 &27.57 &\textbf{4.23} &3.30 &13.58 &4.66 &5.26 &\underline{57.33} &\underline{14.40} &\underline{36.19} &3.26 &39.86 &\underline{7.80} &36.22 &\textbf{24.28} &13.08 &\underline{16.33} &9.52 &4.97 \\
DSOcc-S (Ours) & & 44.88 & 17.51 & 27.39 & 3.05 & 4.38 & 13.62 & 4.82 & \underline{5.75} & 57.23 & 14.38 & 35.46 & 2.74 & \underline{40.85} & 7.46 & 35.89 & 22.90 & 12.15 & 15.75 & 7.71 & 3.72  \\
\bottomrule
\end{tabular}
}

\label{table3}
\end{table*}

\definecolor{barriercol}{RGB}{169,169,169}
\definecolor{bicyclecol}{RGB}{215,25,28}
\definecolor{buscol}{RGB}{253,141,60}
\definecolor{carcol}{RGB}{255,222,0}
\definecolor{convcol}{RGB}{240,90,20}
\definecolor{motorcyclecol}{RGB}{227,26,28}
\definecolor{pedestriancol}{RGB}{31,120,180}
\definecolor{trafficconecol}{RGB}{0,102,102}
\definecolor{trailercol}{RGB}{255,127,0}
\definecolor{truckcol}{RGB}{235,138,98}
\definecolor{drivecol}{RGB}{48,188,172}
\definecolor{sidewalkcol}{RGB}{100,0,100}
\definecolor{terraincol}{RGB}{140,200,100}
\definecolor{manmadecol}{RGB}{200,180,120}
\definecolor{vegetationcol}{RGB}{0,153,0}
\definecolor{otherflatcol}{RGB}{128,128,128}
\definecolor{othercol}{RGB}{0,0,0}
\begin{table*}[htb]
  \centering
  \caption{\textbf{The quantitative results on Occ3D-nuScenes validation set.}\textbf{Bold} and \underline{underline} represent the best and second best results, respectively. Pub. indicates the year of publication and the journal or conference in which the work appeared.}
  \setlength{\tabcolsep}{3pt}
  \resizebox{\textwidth}{!}{%
    \begin{tabular}{l|>{\columncolor{gray!30}}l|ccccccccccccccccc}
      \toprule
      Method  & mIoU
        & \rotatebox{90}{\tikz{\fill[othercol] (0,0) rectangle (0.6em,0.6em);} other}
        & \rotatebox{90}{\tikz{\fill[barriercol] (0,0) rectangle (0.6em,0.6em);} barrier}
        & \rotatebox{90}{\tikz{\fill[bicyclecol] (0,0) rectangle (0.6em,0.6em);} bicycle}
        & \rotatebox{90}{\tikz{\fill[buscol] (0,0) rectangle (0.6em,0.6em);} bus}
        & \rotatebox{90}{\tikz{\fill[carcol] (0,0) rectangle (0.6em,0.6em);} car}
        & \rotatebox{90}{\tikz{\fill[convcol] (0,0) rectangle (0.6em,0.6em);} cons.\ veh.}
        & \rotatebox{90}{\tikz{\fill[motorcyclecol] (0,0) rectangle (0.6em,0.6em);} motorcycle}
        & \rotatebox{90}{\tikz{\fill[pedestriancol] (0,0) rectangle (0.6em,0.6em);} pedestrian}
        & \rotatebox{90}{\tikz{\fill[trafficconecol] (0,0) rectangle (0.6em,0.6em);} traffic cone}
        & \rotatebox{90}{\tikz{\fill[trailercol] (0,0) rectangle (0.6em,0.6em);} trailer}
        & \rotatebox{90}{\tikz{\fill[truckcol] (0,0) rectangle (0.6em,0.6em);} truck}
        & \rotatebox{90}{\tikz{\fill[drivecol] (0,0) rectangle (0.6em,0.6em);} drive.\ surf.}
        & \rotatebox{90}{\tikz{\fill[otherflatcol] (0,0) rectangle (0.6em,0.6em);} other flat}
        & \rotatebox{90}{\tikz{\fill[sidewalkcol] (0,0) rectangle (0.6em,0.6em);} sidewalk}
        & \rotatebox{90}{\tikz{\fill[terraincol] (0,0) rectangle (0.6em,0.6em);} terrain}
        & \rotatebox{90}{\tikz{\fill[manmadecol] (0,0) rectangle (0.6em,0.6em);} manmade}
        & \rotatebox{90}{\tikz{\fill[vegetationcol] (0,0) rectangle (0.6em,0.6em);} vegetation} \\
      \midrule

      MonoScene \cite{cao2022monoscene}        &6.06           & 1.75 & 7.23  & 4.26  & 4.93   & 9.38   &5.67    & 3.98   & 3.01   & 5.90   & 4.45  & 7.17   & 14.91  & 6.32  & 7.92  & 7.43  & 1.01  & 7.65 \\
      OccFormer \cite{zhang2023occformer}      &21.93          &5.94  &30.29  &12.32  & 34.40  & 39.17  &14.44   & 16.45  & 17.22  & 9.27   & 13.90 & 26.36  & 50.99  & 30.96 & 34.66 & 22.73 & 6.76  & 6.97 \\
      TPVFormer \cite{huang2023tri}            &27.83          &7.22  &\underline{38.90}  &13.67  &\textbf{40.78}   &\textbf{45.90}   &\textbf{17.23}   &19.99   &18.85   &14.30   &\textbf{26.69}  & \textbf{34.17}  & 55.65  &\underline{35.47}  & 37.55 &30.70  & 19.40 &16.78 \\
      CTF-Occ \cite{tian2023occ3d}             &\textbf{28.53} &\underline{8.09}  &\textbf{39.33}  &\textbf{20.56}  &38.29   &42.24   &\underline{16.93}   &\textbf{24.52}   &\textbf{22.72}   &\textbf{21.05}   & \underline{22.98} & \underline{31.11}  &53.33   &33.84  &37.98  &33.23  & \underline{20.79} &18.00 \\    
      BEVFormer \cite{li2024bevformer}         &26.88          & 5.85 &37.83  &\underline{17.87}  &\underline{40.44}   &\underline{42.43}   &7.36    &\underline{23.88}   &\underline{21.81}   &\underline{20.98}   &22.38  &30.70   &55.35   &28.36  &36.00   &28.06  &20.04  &17.69 \\
      RenderOcc \cite{pan2024renderocc}        &26.11          &4.84  & 31.72 & 10.72 &27.67   & 26.45  & 13.87  &18.20   &17.67   &17.84   &21.19  &23.25   &\underline{63.20}   &\textbf{ 36.42} & \textbf{46.21} & \underline{44.26} & 19.58 & \underline{20.72} \\ \midrule
      Ours                                     &\underline{28.40} & \textbf{9.05} & 32.07 & 15.22  & 26.51  & 39.66 & 13.65 & 15.35  & 17.62 & 18.40 & 17.64 & 22.53 & \textbf{73.73} & 34.67 & \underline{42.75} & \textbf{45.68} & \textbf{30.53} & \textbf{27.80}  \\
      \bottomrule
    \end{tabular}%
  }
  
  \label{table3_2}
\end{table*}

\begin{figure*}[h] 
\begin{center}
\includegraphics[width=\linewidth]{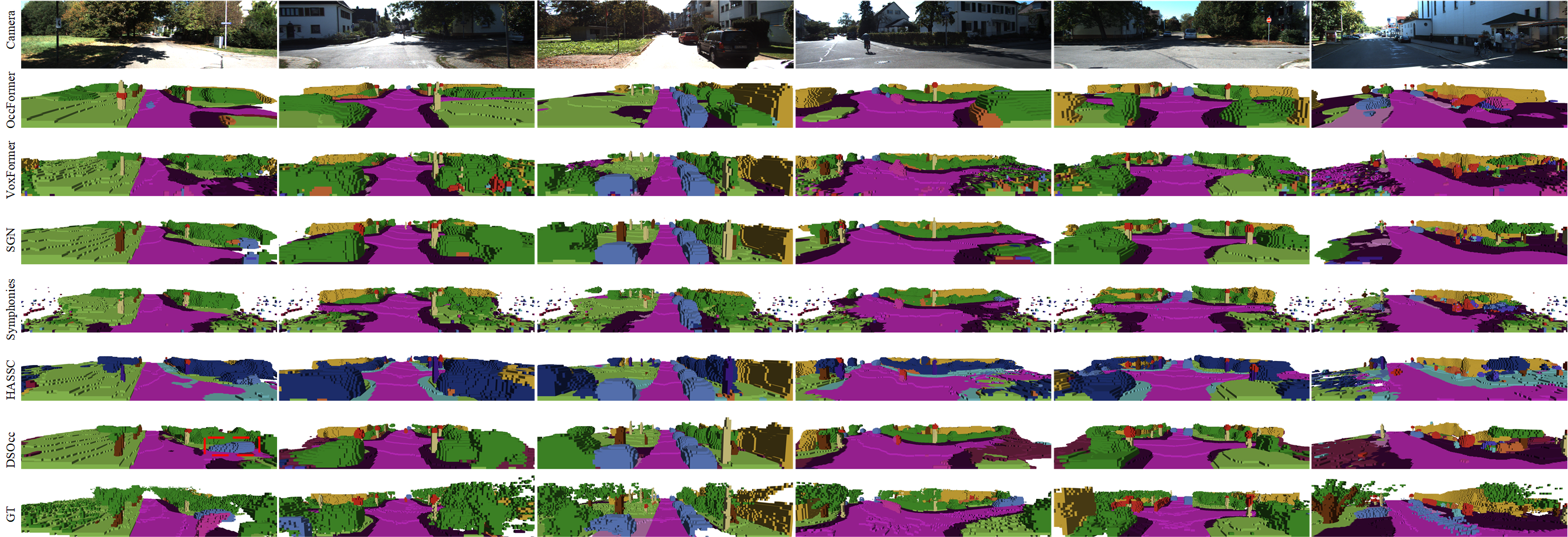}  
\end{center}
\caption{The qualitative comparisons on SemanticKITTI validation set. GT denotes the ground truth.}
\label{fig6}
\end{figure*}

\subsection{Main Results}
\label{sec4.3}

\textbf{Qualitative Comparisons.} We present the qualitative comparison results on the SemanticKITTI validation set, the SemanticKITTI hidden test set, the SSCBench-KITTI-360 test set, Occ3D-nuScenes validation set in Table \ref{table1}, Table \ref{table2}, Table \ref{table3}, and Table \ref{table3_2}, respectively. On the SemanticKITTI dataset, DSOcc-T achieves the highest mIoU of 18.02 (+2.14) and 16.90 (+0.17) on the validation and hidden test sets, respectively, outperforming 12 and 9 classes out of the total 19 classes. DSOcc-S also demonstrates state-of-the-art performance among S-version models, with mIoU scores of 14.59 (+0.04) and 14.96 (+0.95) on the validation and hidden test sets, respectively. On the SSCBench-KITTI-360 test set, DSOcc maintains competitive performance, achieving 17.51 mIoU for S-version, demonstrating the robustness of our method. Because Occ3D-nuScenes targets multi-view inputs, unlike our multi-frame setup, we adapt it by disabling multi-segmentation fusion and using only a single frame during DSOcc training. We employ SurroundDepth \cite{wei2023surrounddepth} for depth estimation. Despite these modifications, DSOcc remains competitive, achieving a mIoU of 28.40 \%.

\textbf{Qualitative Comparisons.} We present the qualitative comparison results on the SemanticKITTI validation set in Fig. \ref{fig6}. DSOcc can perceive small objects far from the ego vehicle, such as bicyclists and poles, and can clearly distinguish every vehicle in a convoy. In complex scenes, such as corners and shops, DSOcc demonstrates excellent discrimination ability within a local region. This highlights its long-range and class-balanced scene perception, enhancing the reliability and robustness of autonomous driving.

\subsection{Ablation Studies}
\label{sec4.4}
All ablation studies were conducted on the SemanticKITTI \texttt{val} set for DSOcc-T. We focus on seven aspects: scene range, input representation, feature processing, depth estimation error impact, soft occupancy confidence calculation, frame design and efficiency, and fusion mechanism.

\textbf{Scene Range.}
Following the scene range setting in \cite{li2023voxformer,mei2024camera, wang2024h2gformer, wang2024not}, we compare our method using two scene ranges: ${25.6m \times 25.6m \times 6.4m}$ (${25.6m}$) and ${12.8m \times 12.8m \times 6.4m}$ (${12.8m}$), as depicted in Table \ref{table4}. DSOcc maintains state-of-the-art performance in both scene ranges. The scene near the ego vehicle is critical for ensuring the safety of autonomous driving. With a 31.31 mIoU at the ${12.8m}$ range, DSOcc outperforms the second-best method, SGN-T, by 5.61, indicating its ability to make more accurate predictions in proximity to the ego vehicle.

\begin{table}[h]
\centering
\caption{Method comparisons in different scene ranges. 25.6${m}$ denote the range of ${25.6m \times 25.6m \times 3.2m}$, and 12.8${m}$ denote the range of ${12.8m \times 12.8m \times 1.6m}$.}
\begin{tabular}{l|ll|ll}
\toprule
\multirow{2}{*}{Method} & \multicolumn{2}{c|}{25.6${m}$} & \multicolumn{2}{c}{12.8${m}$} \\
\cmidrule{2-5}
                                       & IoU        & \cellcolor{gray!30}mIoU      & IoU       & \cellcolor{gray!30}mIoU       \\
\midrule
VoxFormer-T \cite{li2023voxformer}     & 57.69             & \cellcolor{gray!30}18.42              & 65.38              & \cellcolor{gray!30}21.55            \\
H2GFormer-T \cite{wang2024h2gformer}   & 58.51             & \cellcolor{gray!30}20.37              & 67.49              & \cellcolor{gray!30}23.43            \\
HASSC-T \cite{wang2024not}             & 58.01             & \cellcolor{gray!30}20.27              & 66.05              & \cellcolor{gray!30}24.10            \\
SGN-T \cite{mei2024camera}             & \textbf{61.90}    & \cellcolor{gray!30}\underline{22.02}  &\underline{70.61}  & \cellcolor{gray!30}\underline{25.70}  \\
Ours                                   & \underline{61.43} & \cellcolor{gray!30}\textbf{26.48}     &\textbf{73.52}      &\cellcolor{gray!30}\textbf{31.31}      \\
\bottomrule
\end{tabular}
\label{table4}
\end{table}

\textbf{Input Representation.} To evaluate the impact of different input representations, we assess four input types: image features in 2D form (Img. 2D), image features in 3D voxel form (Img. 3D), depth, and semantic segmentation map (Sem.). We use the projection-learning module from \cite{li2023voxformer} for the Img. 2D case. As shown in Table \ref{table5}, Img. 3D improves spatial geometry learning by 0.17 mIoU. Incorporating depth enhances scene completion, yielding a 2.90 IoU gain. We report that depth-aware image feature voxels and semantic-aided voxels achieve 15.00 and 11.16 mIoU, respectively, with standard deviations of 15.09 and 12.76 across all classes. The semantic-aided voxel achieves higher precision on low-proportion classes. For example, the truck class (0.16 \%) improves from 3.99 to 10.28. It shows that semantic-aided voxels promote class-balanced predictions.

\begin{table}[h]
\centering
\caption{Ablation study for input representation. Img. 2D denotes image features in 2D form as in \cite{li2023voxformer}; Img. 3D denotes image features in 3D voxel form; Sem. denotes using semantic segmentation map.}
\begin{tabular}{llll|l>{\columncolor{gray!30}}l}                     
\toprule
Img. 2D    &Img. 3D     &Depth      & Sem.    & IoU    & mIoU \\
\midrule
\checkmark &            &            &           & 34.66  & 10.96 \\
           & \checkmark &            &           & 35.48  & 11.13 \textcolor{gray}{(+0.17)} \\
           & \checkmark & \checkmark &           & 38.38  & 15.00 \textcolor{gray}{(+3.87)}\\
           &            & \checkmark &\checkmark & 29.50  & 11.16 \\
           & \checkmark & \checkmark &\checkmark & \textbf{44.41}  & \textbf{18.02}{} \\
\bottomrule
\end{tabular}
\label{table5}
\end{table}

\textbf{Feature Processing.} To assess depth embedding methods, we compare the baseline against two variants: depth-concatenation and depth-truncation. They appends the depth channel or explicitly set a depth threshold to determine occupancy status. As depicted in Table \ref{table6}, these methods either increase the burden on feature fusion learning or restrict feature collection, decreasing mIoU by 5.65 and 5.63, respectively. To assess the impact of multi-frame processing in establishing semantic-aided voxel, we compare the baseline with two strategies: using only a single frame (w/t multi. sem.) and concatenating multiple frames (w/t sem. concat.). These approaches either miss potential class information or fail to identify the importance of individual frames. 

\begin{table}[h]

\centering
\caption{Ablation study on feature processing. Depth Trunc. denotes using a depth threshold to infer occupancy state; Depth Concat. denotes directly concatenating the depth channel with features; Multi. Sem. denotes the use of multi-frame segmentation maps; and Sem. Concat. denotes directly concatenating the channels of multi-frame segmentation maps.}

\begin{tabular}{l|l>{\columncolor{gray!30}}l}                                  
\toprule
                &IoU         &mIoU  \\
\midrule
w/ Depth Trunc.     & 34.89              & 12.37 \textcolor{gray}{(-5.65)}       \\
w/ Depth Concat.    & 43.31              & 16.37 \textcolor{gray}{(-1.65)}       \\
w/o Multi. Sem.      & \textbf{44.65}      & 16.99 \textcolor{gray}{(-1.03)}        \\
w/ Sem. Concat.     & 43.24              & 17.78 \textcolor{gray}{(-0.24)}                                 \\
Ours             & 44.41     & \textbf{18.02} \\
\bottomrule
\end{tabular}
\label{table6}
\end{table}

\begin{figure}[t] 
\begin{center}
\includegraphics[width=\linewidth]{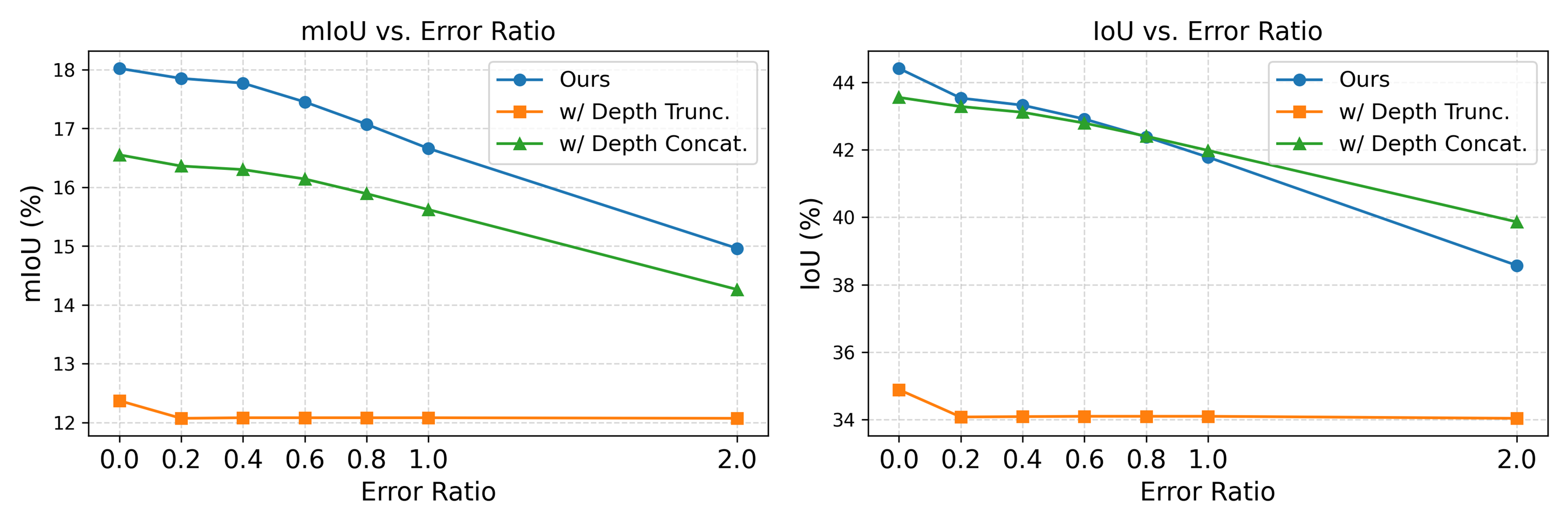}  
\end{center}
\caption{Performance of our method under the varying error ratios of depth estimation.}
\label{figA1}
\end{figure}

\textbf{Depth Estimation Error Impact.}
To assess how depth estimation error affect our model performance, we randomly added noise to the estimated depth to simulate real-world estimation errors. Using the performance of MobileStereoNet \cite{shamsafar2022mobilestereonet} on the KITTI 2015 validation set \cite{geiger2013vision} (EPE = 0.79 px) as the baseline, we tested our method across various error ratios. As illustrated in Fig. \ref{figA1}, our method achieves mIoU scores of 16.06 \% and 14.96 \% when adding one and two times noise, respectively, demonstrating the robustness of soft occupancy confidence.

\textbf{Depth Using.} To assess the advantage of our depth embedding, we compare it against two alternatives: using depth truncation to represent explicit occupancy state and to learn features for occupied voxels \cite{li2023voxformer, wang2024h2gformer}, and using depth to generate geometric cues that interact with image features \cite{jiang2024symphonize,xue2024bi}. As shown in Table \ref{tableA5}, our depth embedding reduces false-negative occupancy predictions and thereby improves mIoU from 13.35 to 15.00. By lifting depth and image features into a spatially aligned 3D representation, our method outperforms the depth-interaction approach, since aligned features enable more accurate fusion. 

\begin{table}[h]
\centering
\caption{Comparison of depth using.}
\begin{tabular}{l|>{\columncolor{gray!30}}ll}
\toprule
                                               & mIoU          & IoU \\ \midrule
Ours                                           &\textbf{15.00} &\textbf{38.38} \\
In  \cite{li2023voxformer, wang2024h2gformer}  &12.37          &34.89 \\
In \cite{jiang2024symphonize,xue2024bi}        & 14.21         & 31.71\\\bottomrule
\end{tabular}
\label{tableA5}
\end{table}

\textbf{Soft Occupancy Confidence Calculation.} 
To compare different methods for computing soft occupancy confidence, we evaluated the performance of tanh and sigmoid functions, each combined with a range-mapping operation to normalize inputs from ${\left( {0,\infty } \right)}$, into ${\left( {1,0} \right)}$. As illustrated in Table \ref{tableA6}, these functions yield comparable results and effectively meet our objectives. However, our exponential-based function achieved a highest mIoU, indicating its superior performance.

\begin{table}[h]
\centering
\caption{Comparison of soft occupancy confidence calculation.}
\begin{tabular}{l|>{\columncolor{gray!30}}ll}
\toprule
        &mIoU           &IoU\\ \midrule
Ours    &\textbf{18.02} &  \textbf{44.41} \\
Tanh    & 14.85         & 42.31  \\
Sigmoid & 14.99         & 39.57  \\
\bottomrule
\end{tabular}
\label{tableA6}
\end{table}

\textbf{Image Segmentation Impact.} To assess the effect of image segmentation on overall performance, we replaced MSeg \cite{lambert2020mseg} with the MobileNet \cite{howard2019searching} and RegSeg \cite{lai2024regseg} trained on the Cityscapes dataset \cite{cordts2016cityscapes}. As illustrated in Table \ref{tableA4}, although these models exhibits lower accuracy and suboptimal class alignment compared to MSeg, it still raises the baseline in mIoU. This improvement confirms the effectiveness of semantic-aided method.

\begin{table}[h]
\centering
\caption{Ablation study for image segmentation impact.}
\begin{tabular}{l|>{\columncolor{gray!30}}ll}
\toprule
                                    & mIoU           & IoU \\ \midrule
Baseline                            &15.00           & 8.38 \\
MSeg                                & \textbf{18.02} & \textbf{144.41}\\
MobileNet\cite{howard2019searching} & 15.01          & 42.67 \\ 
RegSeg\cite{lai2024regseg}          & 15.13          & 41.75\\\bottomrule
\end{tabular}
\label{tableA4}
\end{table}

\textbf{Frame Design and Efficiency.} To evaluate our frame design and efficiency, especially in the final feature fusion, we introduce a comparison called ‘Three Fus’, which fuses lifted image, segmentation, and depth features via dual interaction. Table \ref{tableA7} reports the inference time of our voxel fusion module alongside other voxel-based learning modules. Our depth-aware fusion attains a higher mIoU than ThreeFus. Our voxel fusion processes dense voxels in 177.07 ms, comparable to the sparse-voxel modules used in VoxFormer \cite{li2023voxformer} and SGN \cite{mei2024camera}.

\begin{table}[h]
\centering
\caption{Comparison of fusion mechanisms and efficiency. Three Fus. denotes fusing of lifted
image, segmentation, and depth features via dual interaction.}
\begin{tabular}{l|l|>{\columncolor{gray!30}}ll}
\toprule
                                  &Inf. Time (ms)           & mIoU         & IoU             \\\midrule
Ours                            & 177.07                  &\textbf{18.02} & 44.41              \\
Three Fus.                         &177.07                   & 13.34        & 34.14           \\
In \cite{li2023voxformer}         &84.90                    & 13.35        & 44.15          \\
 In \cite{mei2024camera}          &95.77                    & 15.32        &\textbf{46.21}   \\
 In \cite{jiang2024symphonize}    &\textbf{79.03}           & 13.44        & 41.44           \\
\bottomrule
\end{tabular}

\label{tableA7}
\end{table}

\textbf{Fusion Mechanism.} To evaluate the design of the voxel fusion module, we compare it with five alternatives: using Conv3D layers (Conv.), using Swin Transformer (Swin.), using ${{\bf{\tilde V}}_t^I}$ as the query with \texttt{DA} (Single \texttt{DA} ${Q}$=img.), using ${{\bf{\tilde V}}_t^S}$ as the query with \texttt{DA} (Single \texttt{DA}. ${Q}$=sem.), and using dual \texttt{DA} with 2 layers (\texttt{DA} 2 layers). As depicted in Table \ref{tableA8}, Conv. and Swin. encounter challenges in capturing long-range dependencies, decreasing mIoU by 1.27 and 1.49, respectively. Removing either interaction in baseline also results in  mIoU drop. The 2-layer version does not yield any prediction increase but increases inference time to 283.76. In a nutshell, our voxel fusion module strikes a balance between precision and efficiency.

\begin{table}[H]
\setlength{\tabcolsep}{3pt} 
\caption{Ablation study for fusion mechanism. Inf. Time denotes the inference time of fusion module on per GPUs.}

\centering
\begin{tabular}{l|l|>{\columncolor{gray!30}}ll}                                  
\toprule
                                &Inf. Time (ms)&mIoU               &IoU                        \\
\midrule
Conv.                            &0.16         & 16.75 \textcolor{gray}{(-1.27)} & 40.55              \\
Swin.                            &0.67         & 16.53 \textcolor{gray}{(-1.49)} & 41.30              \\
Single \texttt{DA} (${Q}$=Img.)  &134.52       & 16.94 \textcolor{gray}{(-1.08)} & 44.69              \\
Single \texttt{DA} (${Q}$=Sem.)  &134.12       & 17.88 \textcolor{gray}{(-0.14)} & 42.91              \\
2 Layers \texttt{DA}             &283.76       & 17.48 \textcolor{gray}{(-0.54)} & \textbf{45.16}      \\
Ours                             &177.07       & \textbf{18.02}                  & 44.41              \\
\bottomrule
\end{tabular}

\label{tableA8}
\end{table}

\section{Limitation}
\begin{figure}[!h] 
\begin{center}
\includegraphics[width=\linewidth]{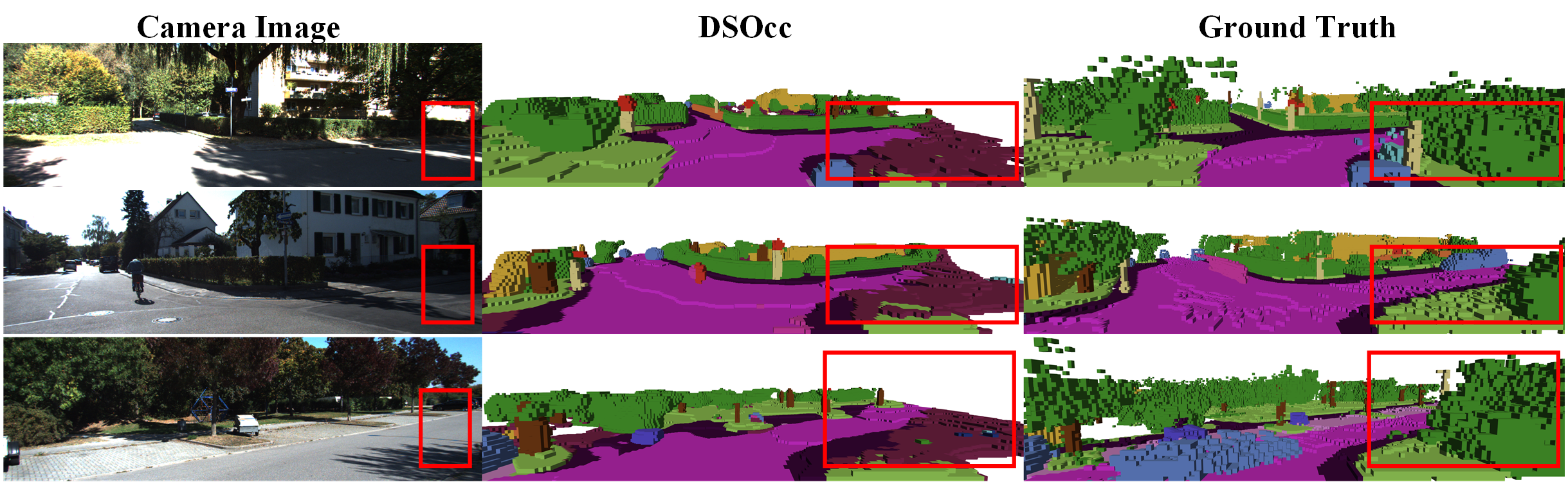}  
\end{center}
\caption{The failure cases.}
\label{figA3}
\end{figure}

As shown in Fig. \ref{figA3}, when the vehicle rounds a corner, our method sometimes fails to reconstruct the side view because of missing camera features and abrupt changes in vehicle calibration. Our method recognizes this region as the flat other-ground class. This issue could be mitigated by incorporating additional camera viewpoints or by integrating other modalities, such as LiDAR.

\section{Conclusion}
In this paper, we propose DSOcc, which leverages depth-aware image feature voxel and semantic-aided voxel to boost camera-based 3D semantic occupancy prediction. DSOcc embeds soft occupancy confidence into image features to enable adaptive implicit occupancy state inference, and utilizes image semantic segmentation to aid occupancy class inference, thereby improving robustness. Experimental results demonstrate that DSOcc achieves state-of-the-art performance on the SemanticKITTI dataset among camera-based methods and achieves competitive performance on the SSCBench-KITTI-360 and Occ3D-nuScenes datasets.

{
\bibliographystyle{IEEEtran}
\bibliography{main}
}

\end{document}